\documentclass{article}

\PassOptionsToPackage{numbers, compress}{natbib}
\usepackage{lmodern}
\usepackage{graphicx}
\usepackage[preprint]{neurips_2024}

\usepackage{enumitem}
\usepackage[utf8]{inputenc} 
\usepackage[T1]{fontenc}    
\usepackage{hyperref}       
\usepackage{url}            
\usepackage{booktabs}       
\usepackage{amsfonts}       
\usepackage{nicefrac}       
\usepackage{microtype}      
\usepackage{xcolor}         
\usepackage{mdframed}
\usepackage{threeparttable}
\usepackage{graphics}
\usepackage{multirow}
\usepackage{rotating}
\usepackage{fontsize}
\usepackage{array}
\usepackage{caption}
\usepackage{xcolor}
\usepackage{graphicx}
\usepackage{colortbl}
\usepackage{listings}
\usepackage{enumitem}
\usepackage{amsmath}
\usepackage{amssymb}
\usepackage{cleveref}
\usepackage{hyperref}
\usepackage{xspace} 
\usepackage{graphicx}
\usepackage{subcaption}

\usepackage[colorinlistoftodos]{todonotes}

\newcommand{\kv}{\texttt{SnapKV}\xspace}

\definecolor{codegreen}{rgb}{0,0.6,0}
\definecolor{codegray}{rgb}{0.5,0.5,0.5}
\definecolor{codepurple}{rgb}{0.58,0,0.82}
\definecolor{backcolour}{rgb}{0.95,0.95,0.92}

\lstdefinestyle{mypython}{
    language=Python,
    basicstyle=\ttfamily\scriptsize,
    keywordstyle=\color{blue},
    stringstyle=\color{red},
    commentstyle=\color{green!70!black},
    backgroundcolor=\color{gray!5},
    showstringspaces=false
}

\lstdefinestyle{mystyle}{
    backgroundcolor=\color{backcolour}, 
    commentstyle=\color{codegreen},
    keywordstyle=\color{magenta},
    numberstyle=\tiny\color{codegray},
    stringstyle=\color{codepurple},
    basicstyle=\fontsize{6pt}{9.6pt}\ttfamily,
    breakatwhitespace=false,         
    breaklines=true,                 
    captionpos=b,                    
    keepspaces=true,                 
    numbers=left,                    
    numbersep=5pt,                  
    showspaces=false,                
    showstringspaces=false,
    showtabs=false,                  
    tabsize=2
}

\title{SnapKV: LLM Knows What You are Looking for Before Generation}

\author{
  Yuhong Li$^{1}$\thanks{equal contribution} \quad Yingbing Huang$^{1}$\footnotemark[1] \quad Bowen Yang$^{2}$ \quad Bharat Venkitesh$^{2}$ \quad Acyr Locatelli$^{2}$ \\
  \textbf{Hanchen Ye$^{1}$ \quad Tianle Cai$^{3}$ \quad Patrick Lewis$^{2}$ \quad Deming Chen$^{1}$} \\
  $^{1}$ University of Illinois Urbana-Champaign \quad $^{2}$ Cohere \quad $^{3}$ Princeton University \\
  \texttt{\small $^{1}$\{leeyh, yh21, hanchen8, dchen\}@illinois.edu} \\
  \texttt{\small $^{2}$\{bowen, bharat, acyr, patrick\}@cohere.com} \\
  \texttt{\small $^{3}$tianle.cai@princeton.edu}
}

\begin{document}

\maketitle
\begin{abstract}
\vspace{-10pt}
Large Language Models (LLMs) have made remarkable progress in processing extensive contexts, with the Key-Value (KV) cache playing a vital role in enhancing their performance. However, the growth of the KV cache in response to increasing input length poses challenges to memory and time efficiency. To address this problem, this paper introduces \kv, an innovative and fine-tuning-free approach that efficiently minimizes KV cache size while still delivering comparable performance in real-world applications.

We discover that each attention head in the model consistently focuses on specific prompt attention features during generation. Meanwhile, this robust pattern can be obtained from an `observation' window located at the end of the prompts. Drawing on this insight, \kv automatically compresses KV caches by selecting clustered important KV positions for each attention head. Our approach significantly reduces the growing computational overhead and memory footprint when processing long input sequences. Specifically, \kv achieves a consistent decoding speed with a 3.6x increase in generation speed and an 8.2x enhancement in memory efficiency compared to the baseline when processing inputs of 16K tokens. At the same time, it maintains comparable performance to the baseline models across 16 long sequence datasets. Moreover, \kv can process up to 380K context tokens on a single A100-80GB GPU using HuggingFace implementation with minor changes, exhibiting only a negligible accuracy drop in the Needle-in-a-Haystack test. Further comprehensive studies suggest \kv's potential for practical applications. 

\end{abstract}
\vspace{-10pt}
\section{Introduction}
\vspace{-5pt}

Many leading LLMs have started to handle longer contexts, overcoming the difficulties in context maintenance and attention mechanism scalability, such as GPT-4~\cite{achiam2023gpt} and Command-R~\cite{coherecommandr} with context length 128K, Claude-3~\cite{anthropic2024claude3} with 200K, and Gemini-Pro-1.5 with 1M~\cite{reid2024gemini}. Despite their impressive capabilities, LLMs still face significant challenges when dealing with long context prompts. Specifically, the KV cache in attention calculation becomes less efficient when processing long context. During inference time, as prompt length increases, the decoding latency per step grows linearly due to the attention calculation across past KVs. Moreover, the large KV cache requires significant memory capacity, increasing hardware demands and limiting model scalability.

There are many approaches to mitigate these problems, such as KV cache eviction during generation stage ~\cite{xiao2023efficient, zhang2024h2o, liu2024scissorhands, ge2023model}. However, most of these methods lack a detailed evaluation in long-context settings. Moreover, they mainly focus on compressing the KV cache appended during decoding steps, while overlooking the realistic problem of compressing KV cache for prompts, which is typically the bottleneck in memory efficiency. 
In practical applications like chatbots and agents, where prompts range from multi-turn conversations to extensive articles or codebases~\cite{achiam2023gpt, liu2021lifelong, bairi2023codeplan}, prompts are often much larger than generated responses such as summaries and code pieces, thus creating significant inference latency and memory utilization overhead. Additional challenge lies in compressing KV cache for such vast prompts without losing crucial information for accurate generation, especially in scenarios with various noisy contexts.

\begin{figure}[t]
    \centering
    \includegraphics[width=\textwidth]{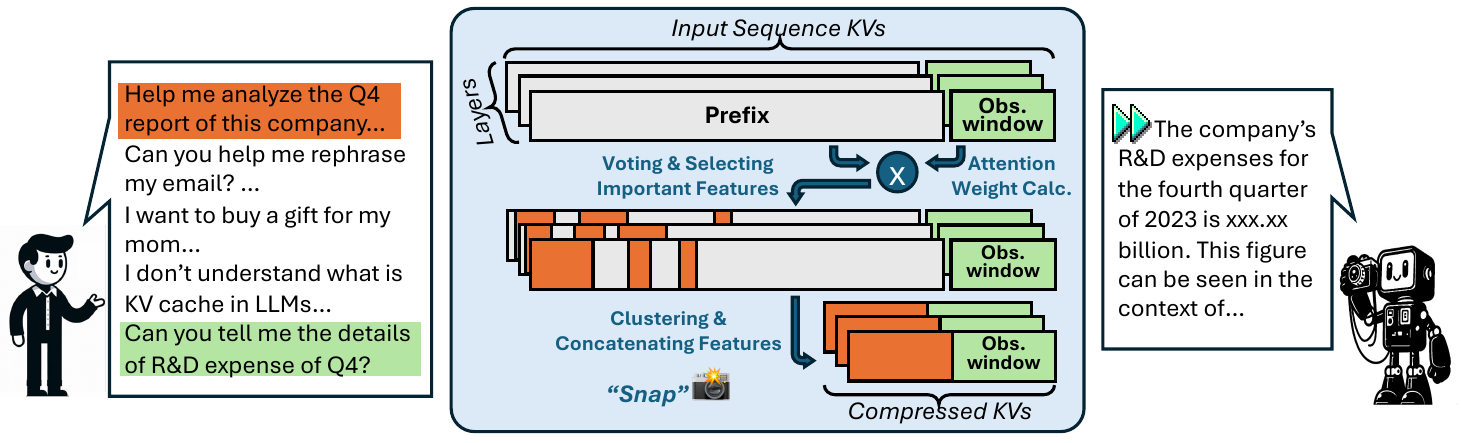}
    \caption{The graph shows the simplified workflow of \kv, where the orange area represents the cluster of features per head selected by \kv. These features are then used to form new Key-Value pairs concatenated with the features in the observation window. Together, the selected prefix and observation windows constitute the new KV cache utilized for the generation.}
    \vspace{-10pt}
    \label{fig: algo}
\end{figure}

In our paper, we find an important attention allocation phenomenon: only a portion of prompt tokens convey essential information for response generation, and these tokens remain unchanged during generation. To validate the robustness of this finding, we design a thorough set of experiments across diverse prompts in terms of length, format, and content. From our observations, we derive an innovative and intuitive method, \kv, which can smartly identify the attention allocation pattern and compress the KV cache for long sequence prompts without compromising the model's accuracy. With its comprehensive design, \kv demonstrates its effectiveness on various datasets and can be easily integrated into popular deep-learning frameworks with just a few code adjustments. Our contributions are as follows:
\begin{itemize}[topsep=0pt,parsep=4pt,itemsep=0pt, leftmargin=16pt]
    \item We design experiments to explore the attention allocation pattern during generation, focusing on two key questions: 
    \begin{enumerate}[topsep=0pt,parsep=0pt,itemsep=0pt, leftmargin=16pt]
        \item Is there a consistent attention allocation pattern for input sequence tokens?
        \item Is it feasible to identify this pattern prior to the generation stage?
    \end{enumerate}
    Our finding suggests that for LLMs, the attention allocation of most input sequence tokens stay consistent during generation. Thus, \textit{LLMs knows what you are looking for before generation}.
    \item Inspired by our observations above, we develop an efficient and fine-tuning-free algorithm, \kv, which efficiently identifies critical attention features and compresses KV cache correspondingly with minimal model modification (See Fig.~\ref{fig: algo}). 

    \item We evaluate \kv across diverse LLMs and long-sequence datasets. \kv shows comparable accuracy with full KV caching method while achieving improved decoding speed and memory efficiency. Meanwhile, we conduct the pressure test with Needle-in-a-Haystack to further demonstrate its memory efficiency and information retrieval ability.   
\end{itemize}

\vspace{-5pt}
\section{Related Works}
\vspace{-5pt}

Many previous works compress the KV cache by selectively dropping KVs using different algorithms. In StreamLLM~\cite{xiao2023efficient}, only the most recent tokens and attention sinks (first few tokens) are retained to reduce the KV cache size, making it lose the important information carried by the discarded middle tokens \footnote{\url{https://github.com/mit-han-lab/streaming-llm?tab=readme-ov-file\#faq}}. Heavy-Hitter Oracle (H2O)~\cite{zhang2024h2o} introduces a policy that greedily drops KVs during generation based on a scoring function derived from cumulative attention. While this approach effectively compresses the KVs appended to the cache during generation, it overlooks compression of prompt KVs, which is crucial for reducing memory and computational overhead. Building on a similar concept, Adaptive KV Compression (FastGen)~\cite{ge2023model} implements a dual-phase algorithm that encompasses four KV cache compression policies. Initially, it identifies optimal policies through profiling results obtained from prompt encoding. Subsequently, it dynamically evicts caches during the generation phase based on these policies. Nonetheless, it faces the similar problem with H2O. ScissorHands~\cite{liu2024scissorhands} focuses on identifying and retaining pivotal tokens that exhibit a consistent attention weight pattern with previous token windows during generation steps. However, this method concentrates solely on the window of previous pivotal tokens in generation and neglects the extensive prompt that contains essential information for generating accurate responses. This oversight could lead to an inability to extract detailed information from prompts.

In summary, existing methods have not effectively addressed the challenges encountered in real-world applications, where prompts are exceptionally long yet require accurate information retrieval. Although these techniques may reduce the KV cache size during generation, they do not address the primary challenges of understanding complex prompt contexts, leaving critical issues unresolved.

\begin{figure}[t]
    \begin{minipage}[b]{0.48\textwidth}
        \centering
        \includegraphics[width=1.1\textwidth]{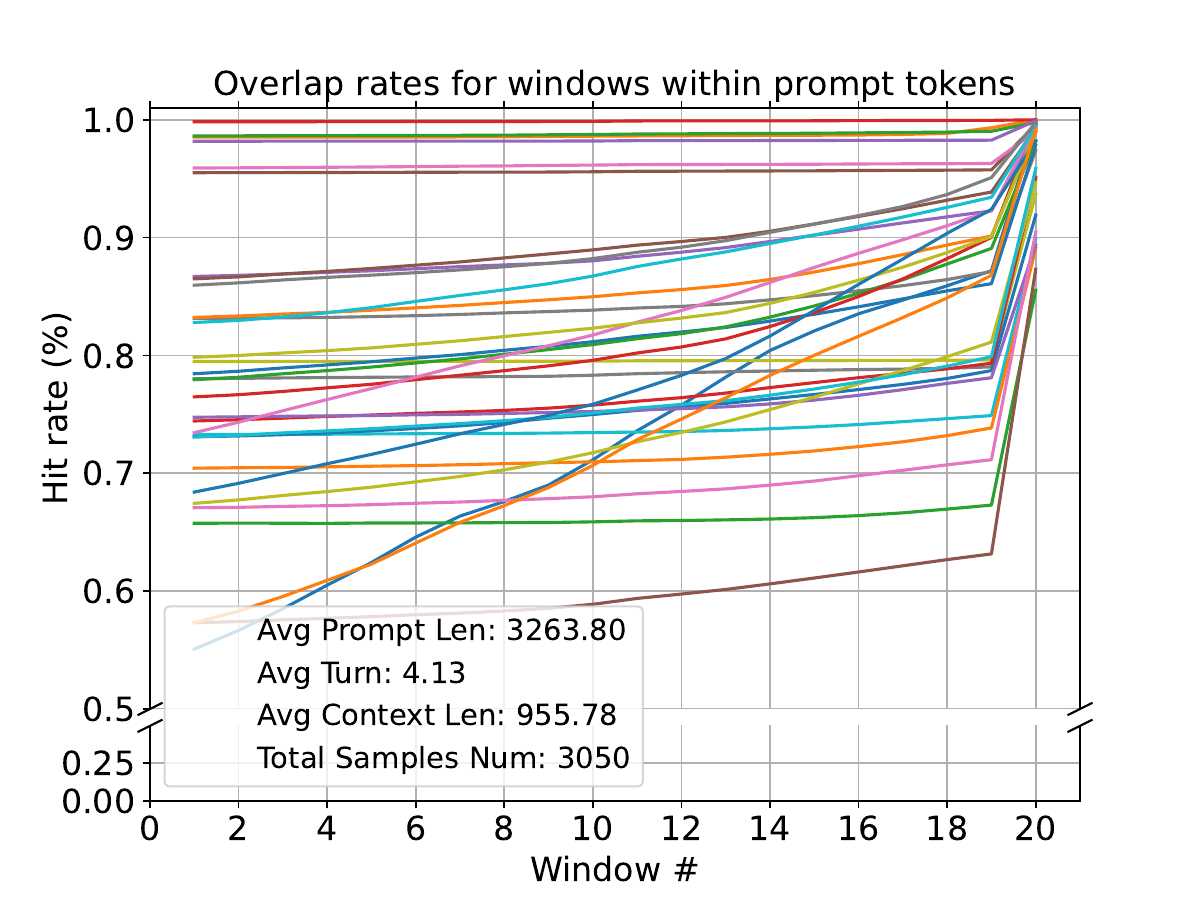}
        \caption{The overlap rates between attention features of the input sequence, selected by various windows along the input and during generation, with each line representing a model layer.}
        \label{fig: hit_rate_prompt}
    \end{minipage}
    \hfill
    \begin{minipage}[b]{0.48\textwidth}
        \centering
        \includegraphics[width=1.1\textwidth]{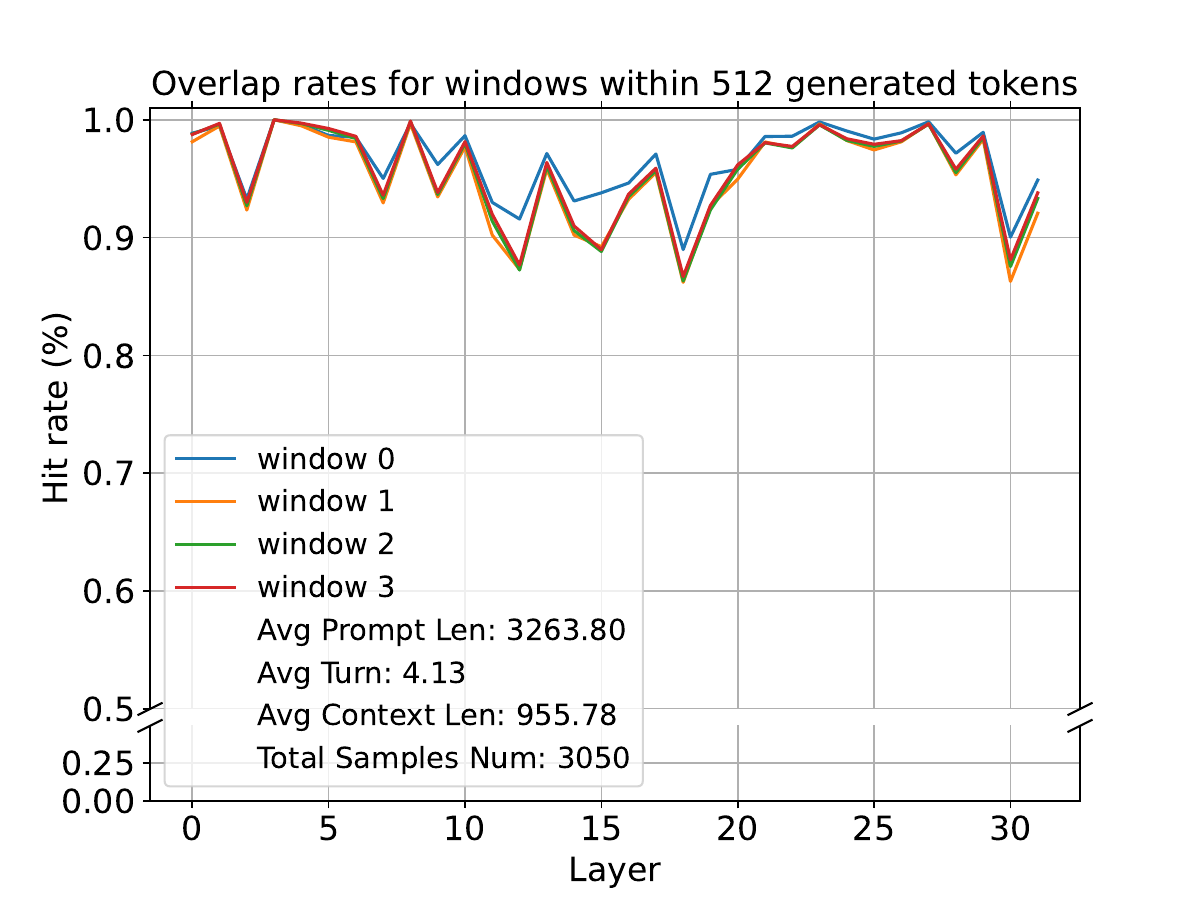}
        \caption{ The layer-wise overlap rates between input sequence attention features selected by the last window of input sequence and those selected by 4 windows along generation.}
        \label{fig: hit_rate}
    \end{minipage}
    \vspace{-10pt}
\end{figure}

\vspace{-5pt}
\section{Observations}
\vspace{-5pt}

In this section, we present our observations regarding the attention allocation patterns in the Query-Key matrix during token generation. Our analysis utilizes samples from Ultrachat~\cite{ding2023enhancing}, a multi-turns, high-quality instruction dataset consisting of 1.4 million dialogues. We further filter the sequences with response length greater than 512 and prompt length greater than 3k. Our findings are concluded into two key observations as follows:

\begin{itemize}[topsep=0pt,parsep=4pt,itemsep=0pt,leftmargin=15pt]
\item \textbf{Pattern can be identified before generation.} In this experiment, we split the attention features of input sequence of each layer into multiple windows, each with 128 tokens, and calculate the averaged attention weights of the last 20 windows separately. To understand the attention allocation patterns along input sequences, we calculate the overlap rates between \emph{important} attention features of input sequence (those with high average attention weights) identified by each window and the actual ones used by generation. The experimental results are shown in Fig.~\ref{fig: hit_rate_prompt}. We observe that the last window of input sequence recognizes highly similar attention allocation pattern with the actual generation.

\item \textbf{Pattern is consistent during generation.} 
We study if the positions of features identified as crucial in the last window of input sequence maintain their significance in the subsequent token generation.  In the experiment, we split the generated tokens into 4 windows for every layer, each spanning 128 tokens, to compute the averaged overlap rates of these windows versus the last window of input sequence. As shown in Fig.~\ref{fig: hit_rate}, active attention features of input sequence obtained from the last window exhibit remarkable consistency throughout the generation process, as evidenced by high overlap rates.
\end{itemize}
\label{sec: consistency_obs}

\vspace{-5pt}
\section{SnapKV}
\label{sec: 4}
\vspace{-5pt}

In the attention mechanism, the growth in prompts will significantly increase time complexity for generation due to the Query-Key matrix multiplication. \kv addresses this issue by maintaining a constant amount of prompt KVs during generation, significantly reducing serving times for long-context LLMs. To structure our method coherently, we propose the following terminologies:

\begin{itemize}[topsep=0pt,parsep=4pt,itemsep=0pt,leftmargin=15pt]
\item \textbf{Prompt Length (}\(L_{\text{prompt}}\)\textbf{):} The total length of the user-provided input.

\item \textbf{Observation Window (}\(L_{\text{obs}}\)\textbf{):} The last segment of the prompt. This window is crucial for analyzing the influence of different contexts on attention allocation patterns.

\item \textbf{Prefix Length (}\(L_{\text{prefix}}\)\textbf{):} The length of the input preceding the observation window. It is part of the prompt and does not include the observation window. Overall, we have:
\begin{small}
\begin{equation}\label{eq: length}
L_{\text{prompt}} = L_{\text{prefix}} + L_{\text{obs}}
\end{equation}
\end{small}

\item \textbf{Voting:} The process of calculating attention weights for each query within the observation window across all heads, aggregating these weights to highlight the prefix positions that are considered most significant. For a single batch of sequence, formally:
\begin{small}
\begin{align}
\label{eq: voting}
\mathbf{C} = \sum_{i=0}^{L_{\text{obs}}} \mathbf{W}_{\text{obs}}[:, i, :]\\
\quad I = \text{Top}_k(\mathbf{C}, k)
\end{align}
\end{small}
where \(\text{Top}_k(\mathbf{C}, k)\) selects the indices $I$ of the top \(k\) values in tensor \(\mathbf{C}\) per head. \(k\) is defined as \(\left\lfloor\textit{p} \times L_{\text{prefix}}\right\rfloor\), where $p$ stands for the compression rate. The tensor \(\mathbf{W}_{\text{obs}}\in \mathbb{R}^{N \times L_{\text{obs}} \times L_{\text{prefix}}}\) represents the subset of the prompt softmax-normalized attention features over $N$ heads.

\item \textbf{Hit Rate:} We define attention features above a predefined threshold \(\theta\) during generation as \emph{important} features. The hit rate, \( H \), is the number of important features successfully selected by the previous voting process over the total number of important features. \( H \) quantifies the effectiveness of the voting mechanism and is calculated as follows:
\begin{small}
\begin{align}
\mathbf{M}_{\text{vote\_obs}} &= \text{zeros\_like} (\mathbf{A}_{\text{cur}}) \\
\mathbf{M}_{\text{vote\_obs}}[I] &= 1 \\
\mathbf{M}_{\text{threshold\_cur}} &= \mathbf{1} (\mathbf{A}_{\text{cur}} > \theta) \\
\mathbf{O} &= \mathbf{M}_{\text{threshold\_cur}} \land \mathbf{M}_{\text{vote\_obs}} \label{eq:stepa} \\
H &= \frac{\sum \mathbf{O}}{\sum \mathbf{M}_{\text{threshold\_cur}}} \label{eq:stepb}
\end{align}
\end{small}
\(\mathbf{A}_{\text{cur}}\in \mathbb{R}^{N \times L_{\text{prefix}}}\) represents the attention features between the current generated query and prefix keys. \(\mathbf{M}\) selects attention features by indices. The threshold operation filters \(\mathbf{A}_{\text{cur}}\) to retain only features with values over \(\theta\), indicating important attention activations. The \(\mathbf{O}\) measures the overlap between attention features selected by \(\mathbf{M}_{\text{threshold\_cur}}\) and \(\mathbf{M}_{\text{vote\_obs}}\), quantifying the alignment of the current attention with previously identified important features. The hit rate \(H\) is then computed as the ratio of the sum of overlap \(\mathbf{O}\) to the sum of important features \(\mathbf{M}_{\text{threshold\_cur}}\), providing a metric for the efficacy of the attention mechanism in recognizing and emphasizing important attention features within the context. We use \(\mathcal{H}(\mathbf{M}_{\text{threshold\_cur}}, \mathbf{M}_{\text{vote\_obs}})\) to denote combination of Eq.~\ref{eq:stepa} and Eq.~\ref{eq:stepb}.
\end{itemize}

\vspace{-5pt}
\subsection{Observation Window-based Algorithm}
\vspace{-5pt}

The core approach of \kv involves identifying and selecting the most crucial attention features per head to create the compressed KV cache. Listing~\ref{lst:pytorch} shows the PyTorch-style pseudo code of \kv. Overall, \kv operates through two stages as follows:

\begin{itemize}[topsep=0pt,parsep=4pt,itemsep=0pt,leftmargin=15pt]
    \item \textbf{Vote for important previous features.}
    By the voting process defined above (Eq.~\ref{eq: voting}), we select the important attention features based on the observation window. Sec.~\ref{sec: consistency_obs} highlights the consistency of the attention allocation pattern within observation windows throughout the generation, suggesting that these selected attention features are also vital for subsequent generation. Furthermore, we implement clustering to retain the features surrounding the selected attention features (Sec. \ref{sec: clustering}). Line~\ref{line:start_vote}-\ref{line:end_vote} shows the pseudo code of the voting process.
    
    \item \textbf{Update and store compressed keys and values.}
    We concatenate the selected attention features with all features within the observation window, which encompasses all features containing the necessary prompt information. Line~\ref{line:start_compress}-
    \ref{line:end_compress} shows the compressing process. The concatenated KVs are stored for later use in generation, thereby saving memory usage.
\end{itemize}

\begin{lstlisting}[language=Python, style=mypython, escapechar=|, caption=Implementation of \kv in pseudo PyTorch style., label={lst:pytorch}]
def snap_kv(query_states, key_states, value_states, window_size, max_capacity_prompt, kernel_size):
    bsz, num_heads, q_len, head_dim = query_states.shape
    # Ensure it is the prompt phase.
    assert key_states.shape[-2] == query_states.shape[-2]
    if q_len < max_capacity_prompt:
        return key_states, value_states
    else:
        # Compute attention weights of observing window's queries and prefix context's Keys. |\label{line:start_vote}|
        attn_weights = compute_attn(query_states[..., -window_size:, :], key_states, attention_mask) 
        # Sum the weight along the query dimension.
        vote = attn_weights[..., -window_size:, :-window_size].sum(dim=-2)
        # Apply 1D pooling for clustering.
        pool_vote = pool1d(vote, kernel_size=kernel_size, padding=kernel_size//2, stride=1) |\label{line:pooling}|
        # Select top-k indices based on the pooled weights to identify important positions.
        indices = pool_vote.topk(max_capacity_prompt - window_size, dim=-1).indices
        # Expand the indices to match the head dimension for gathering.
        indices = indices.unsqueeze(-1).expand(-1, -1, -1, head_dim) |\label{line:end_vote}|
        # Gather the compressed past key and value states based on the selected indices. |\label{line:start_compress}|
        k_past_compress = key_states[..., :-window_size, :].gather(dim=2, index=indices)
        v_past_compress = value_states[..., :-window_size, :].gather(dim=2, index=indices)
        k_obs = key_states[..., -window_size:, :]
        v_obs = value_states[..., -window_size:, :]
        key_states = torch.cat([k_past_compress, k_obs], dim=2)
        value_states = torch.cat([v_past_compress, v_obs], dim=2) |\label{line:end_compress}|
        return key_states, value_states
\end{lstlisting}
\vspace{-5pt}

\vspace{-5pt}
\subsection{Robustness Analysis of Hit Rate}
\vspace{-5pt}

To understand the robustness of the observation window-based algorithm, we analyze its hit rate on multiple long documents QA datasets including QMSum \cite{zhong2021qmsum}, a query-based multi-domain meeting summarization; Openreview \cite{an2023eval}, a collection of papers from \texttt{openreview.net}; SPACE \cite{angelidis2021extractive}, an extractive opinion summarization in quantized transformer spaces.
The model we probe is \texttt{Mistral-7B-Instruct-v0.2}. Overall, we want to answer the following two questions:
\begin{enumerate}[topsep=0pt,parsep=0pt,itemsep=0pt,leftmargin=15pt]
    \item Does the nature of instructions in the prompt affect the hit rate?
    \item Does the context and instruction positioning affect the hit rate?
\end{enumerate}

\vspace{-5pt}
\subsubsection{Contextual Dependency of Patterns}
\vspace{-5pt}

\begin{figure}[t]
    \centering
    \includegraphics[width=0.7\textwidth]{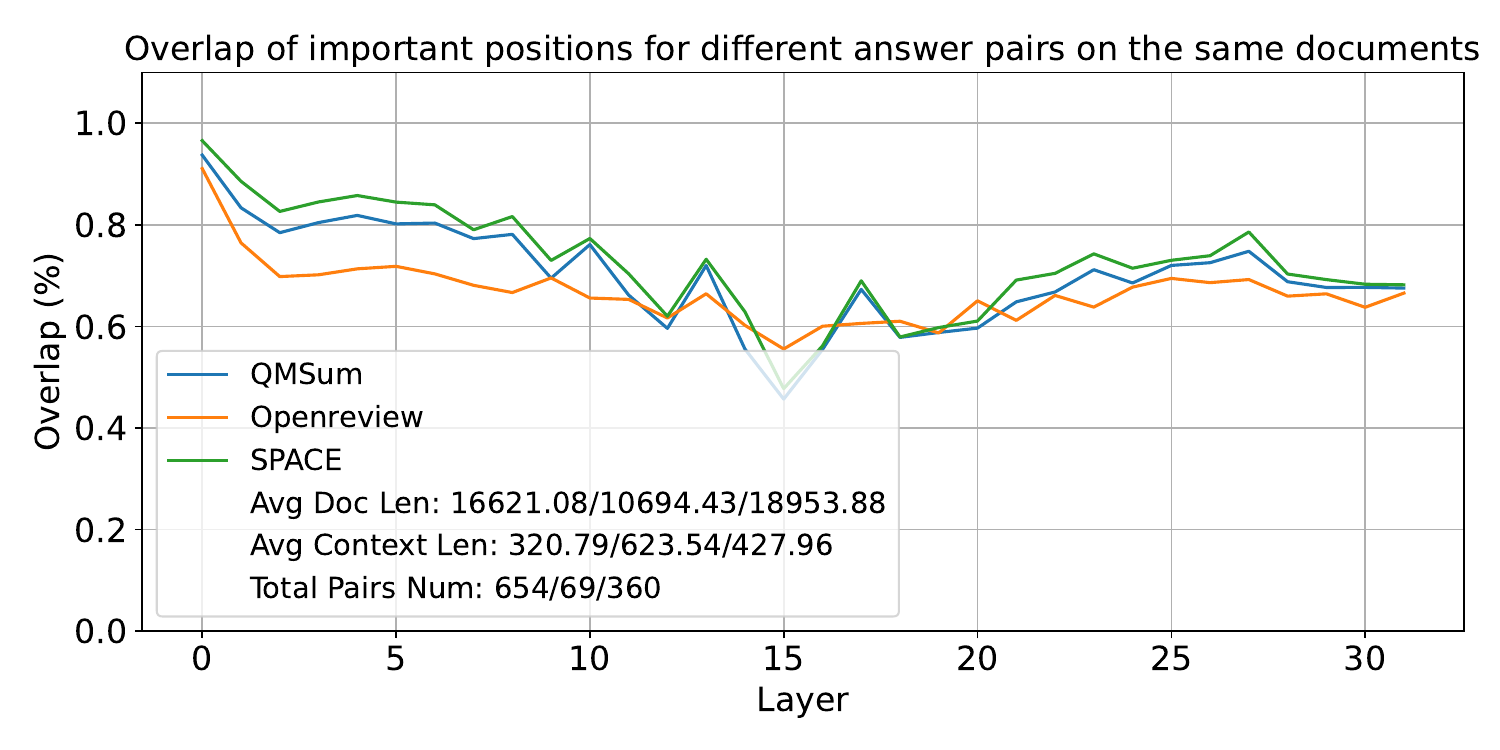}
    \vspace{-10pt}
    \caption{The layer-wise overlap of important positions utilized by different question-answer pairs in the same dataset.}
    \vspace{-10pt}
    \label{fig: qa_pairs}
\end{figure}

We analyze whether instructions will affect the selection of important features even if the provided context is the same. Our experiment utilizes different instructions on the same document and selects the important features based on the observation window that consists of both the instructions and their corresponding responses. Then we calculate the hit rates between important features selected by different instruction-response pairs within the same document by using \(\mathcal{H}(\text{M}_{\text{vote\_A}}, \text{M}_{\text{vote\_B}})\). By varying the instructions, we observe that different instructions prioritize different prefix attention features, as indicated by the descending trend in hit rates shown in Fig. \ref{fig: qa_pairs}. Our findings reveal an interesting aspect of KV cache in LLMs: the important attention features change with different instructions. This variability challenges the effectiveness of static compression methods that depend on constant weighted importance or fixed policies~\cite{liu2024scissorhands,zhang2024h2o,ge2023model}. Thus, the complex relationship between context and related KV cache emphasizes the need for context-aware compression strategies and highlights the capability of \kv that recognizes this dynamic.

\vspace{-5pt}
\subsubsection{Invariance to Instruction Positions}
\vspace{-5pt}

\begin{figure}[t]
    \centering
    \includegraphics[width=0.49\textwidth]{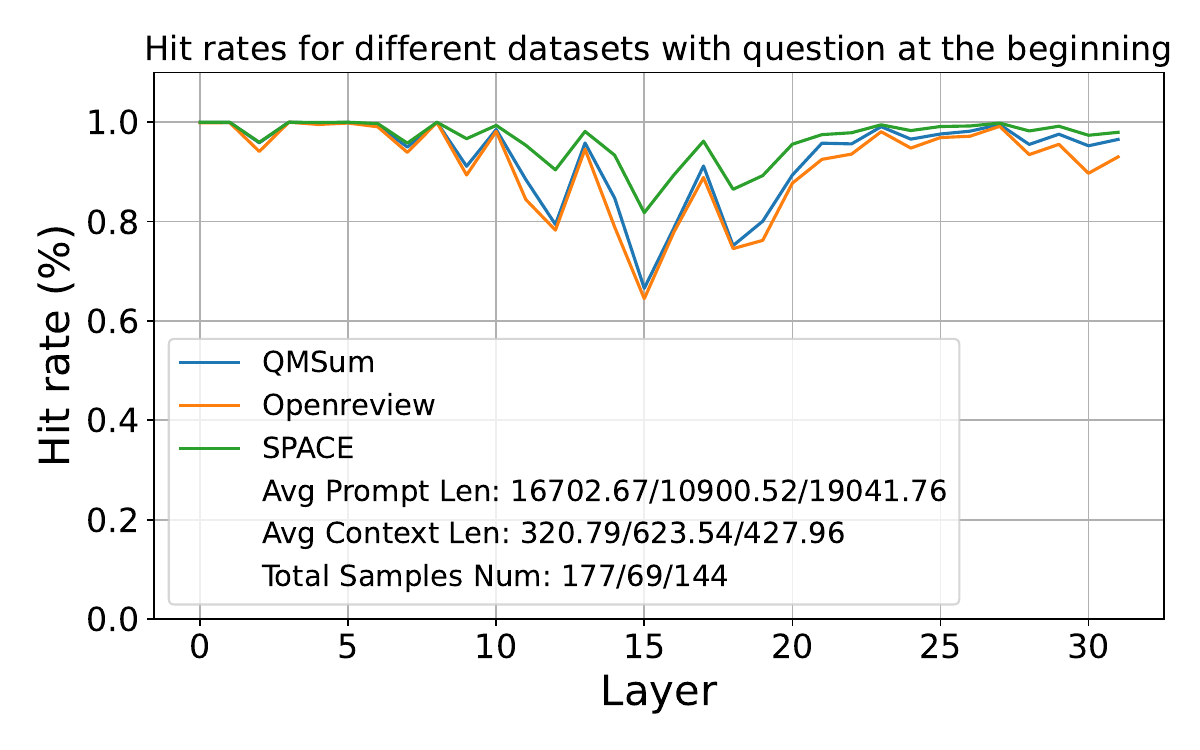}
    \includegraphics[width=0.49\textwidth]{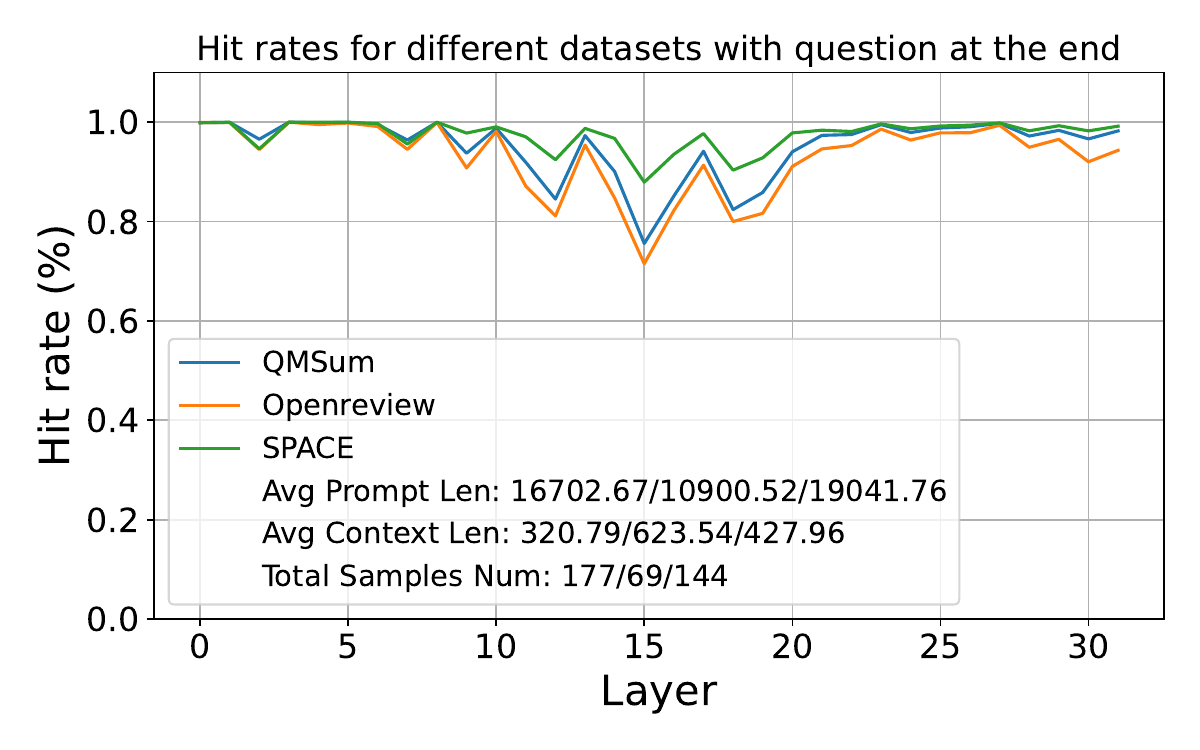}
    \vspace{-10pt}
    \caption{The layer-wise average hit rate of important positions used by prompts with questions at the beginning and the end.}
    \vspace{-10pt}
    \label{fig: question_pos}
\end{figure}

Our analysis also extends to the significance of instruction positioning on the interpretability of LLMs and their selection of important features. We calculate the average hit rate for the responses using the same observation window size as in the previous experiment. Our results shown in Fig. \ref{fig: question_pos} indicate that across all three datasets, the hit rates are consistently high regardless of whether instructions are positioned before or after extensive supplementary contexts. This consistency suggests that \kv is able to identify attention allocation patterns regardless of the question's positions.

\vspace{-5pt}
\subsection{Efficient Clustering via Pooling}
\label{sec: clustering}
\vspace{-5pt}

In LLMs, information retrieval and generation rely on features with high attention weight and are supplemented by copying the rest of features in context using induction heads~\cite{olsson2022context}. Hence, naively selecting the top features results in retaining only portions of details and then losing the completeness of the information. For example, such compression might cause the LLMs to retrieve only the country code of a phone number and hallucinate the rest. Our experiment also revealed that only selecting the features with the highest weights is insufficient (Sec.~\ref{sec:ablation}). Such sparse selection risks compromising the contextual integrity encapsulated in between features, thereby reducing accuracy. Based on the insights, we propose a fine-grained clustering algorithm utilizing a pooling layer shown in Line~\ref{line:pooling}.

\vspace{-5pt}
\section{Experiments}
\vspace{-5pt}
 \label{sec: experimentes}

In our experimental setup, we explore the performance of \kv across models that can handle extended prompt sequence contexts. First, we deliver a pressure test and benchmark the speed of \texttt{LWM-Text-Chat-1M}~\cite{liu2024world}, which is state-of-the-art regarding its context length. 
We then conduct an ablation study on \texttt{Mistral-7B-Instruct-v0.2} to understand the influence of pooling on the model's information retrieval performance. We assess model performances using the LongBench~\cite{bai2023longbench} dataset. Further, we dive into a comprehensive examination of the \texttt{Command-R}~\cite{coherecommandr} model, another leading open-source model in the field. Lastly, we show that \kv can be utilized with other acceleration strategies such as parallel decoding.

\vspace{-5pt}
\subsection{Benchmarks on LWM-Text-Chat-1M}
\vspace{-5pt}

\texttt{LWM-Text-Chat-1M}~\cite{liu2024world} is a 7B instruction-fine-tuned model with up to one million context length. In this section, we conduct a pressure test on this model and examine its algorithmic efficiencies.

\begin{figure}[!t]
    \centering
    \vspace{-5pt}
    \includegraphics[width=0.95\textwidth]{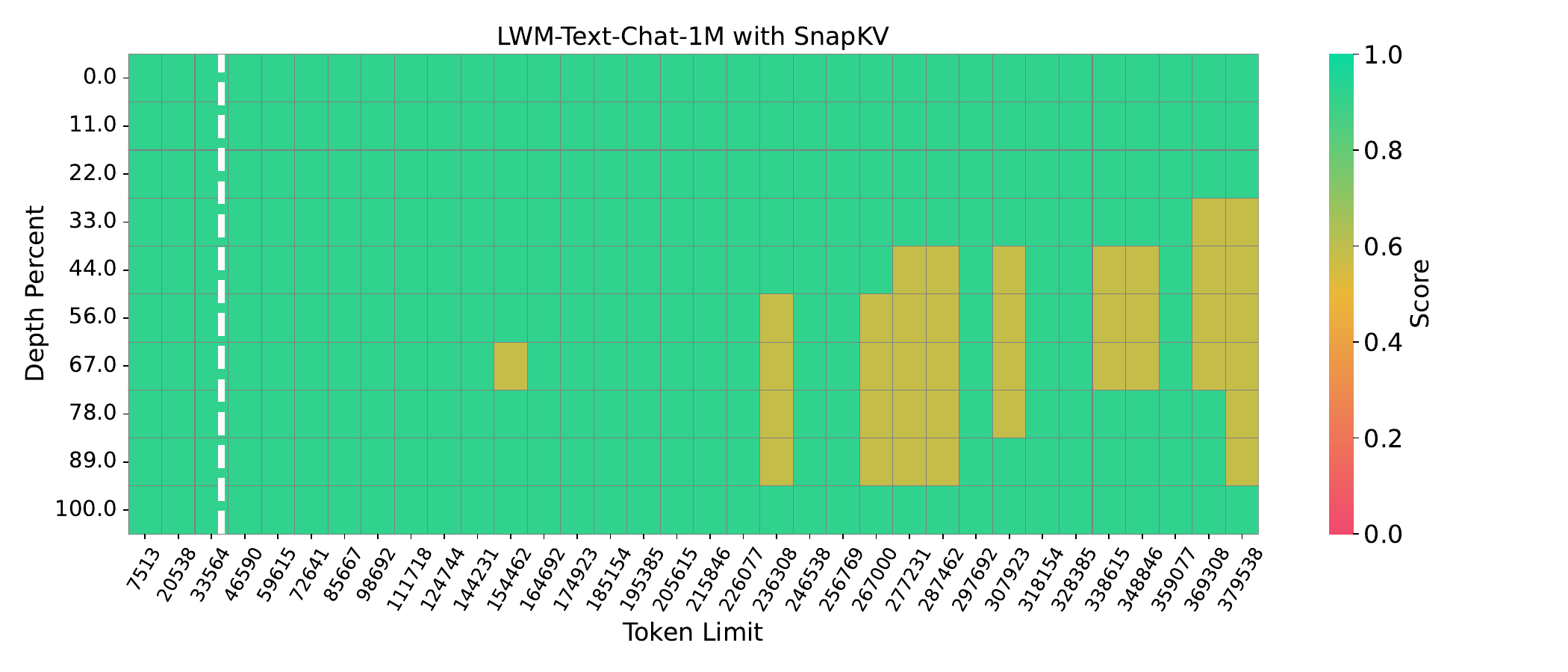}
    \vspace{-5pt}
    \caption{Needle-in-a-Haystack test performance comparison on single A100-80GB GPU, native HuggingFace implementation with only a few lines of code changed. The x-axis denotes the length of the document (the “haystack”) from 1K to 380K tokens; the y-axis indicates the position that the “needle” (a short sentence) is located within the document. For example, 50\% indicates that the needle is placed in the middle of the document. Here LWMChat with \kv is able to retrieve the needle correctly before 140k and with only a little accuracy drop after. Meanwhile, the original implementation encounters OOM error with 33k input tokens (white dashed line).}
    \vspace{-10pt}
    \label{fig: needle}
\end{figure}

\vspace{-5pt}
\subsubsection{Needle-in-a-Haystack} 
\vspace{-5pt}

The Needle-in-a-Haystack test \cite{kamradt2023needle} challenges the model to accurately retrieve information from a specific sentence ("needle") concealed within an extensive document (the "haystack"), with the sentence placed at a random location. Typically, sentences that are inserted in the middle of prompts are harder to retrieve.
To rigorously evaluate \kv's capabilities, we extended the document length to 380k tokens which is the longest content that can be processed by a single A100-80GB GPU. We configured the prompt KV cache size to 1024, enabling \kv to select the most crucial 1024 attention features from the prompt for answer generation, with a maximum pooling kernel size of 5 and an observation window size of 16, both of which are hyperparameters that can be customized. The compelling outcomes in Fig. \ref{fig: needle} from the Needle-in-a-Haystack test underscore \kv's potential to precisely manage small details on extremely long input contexts with a 380x compression ratio. 

\begin{figure}[!t]
    \centering
    \includegraphics[width=0.95\textwidth]{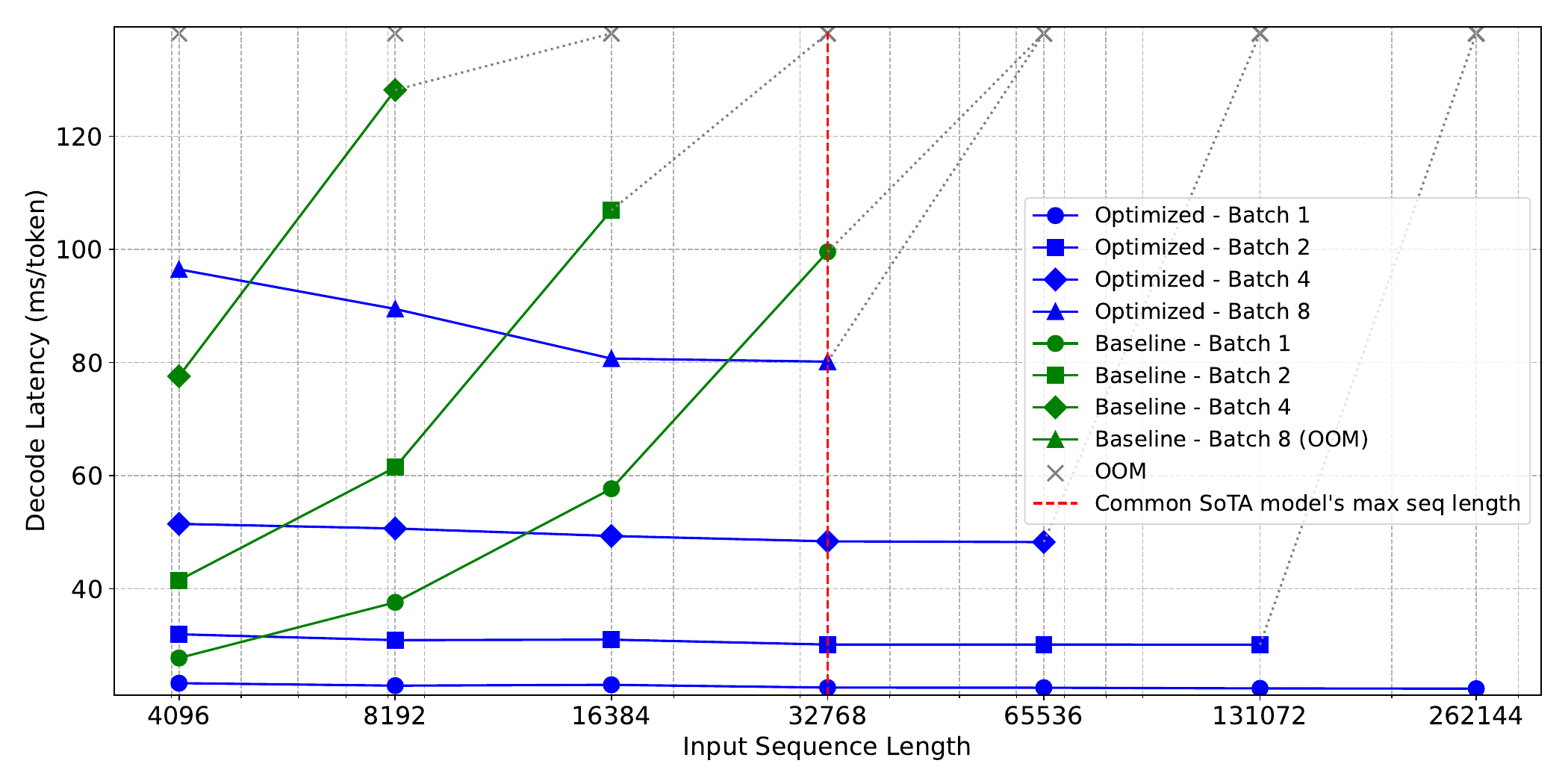}
    \vspace{-5pt}
    \caption{Decoding latency comparison of baseline implementation and \kv optimized solutions on various batch sizes. The x-axis denotes the input sequence length; the y-axis indicates decoding latency (ms/token). All experiments are conducted on an A100 80GB GPU. The red dotted line denotes the common context length of state-of-the-art long sequence models.}
    \vspace{-5pt}
    \label{fig: speed}
\end{figure}

\vspace{-5pt}
\subsubsection{Decoding Speed and Memory Bound}
\vspace{-5pt}

We further benchmark the speed of \texttt{LWM-Text-Chat-1M} under different batch-size settings using \kv. We set the maximum KV cache size as 2048 for \kv, and fix the generation length at 512 to ensure a fair comparison. There are two main takeaways from our experiment on decoding speed and prompt sequence length on various batch sizes, as shown in Fig.~\ref{fig: speed}. First, as the input sequence length increases, the decoding latency of the baseline implementation escalates linearly. Conversely, the \kv-optimized model maintains a constant decoding speed since the compressed KV cache size of prompt stays the same regardless of input sequence length and there is no extra update during the inference. For instance, at a sequence length of 16k and a batch size of 2, the decoding time for the baseline model surpasses 100 ms, whereas for \kv-optimized model, the decoding time consistently remains below 40 ms, achieving approximately a 3.6x speedup. Second, with the same batch size, the model integrated with \kv can decode significantly longer sequences. For example, at a batch size of 2, the baseline model encounters an OOM error beyond 16k input tokens, whereas the \kv-enhanced model extends this limit to 131k input tokens, indicating an approximately 8.2x improvement. This demonstrates \kv's effectiveness in minimizing memory consumption.



\begin{figure}[!t]
    \centering
    \includegraphics[width=0.49\textwidth]{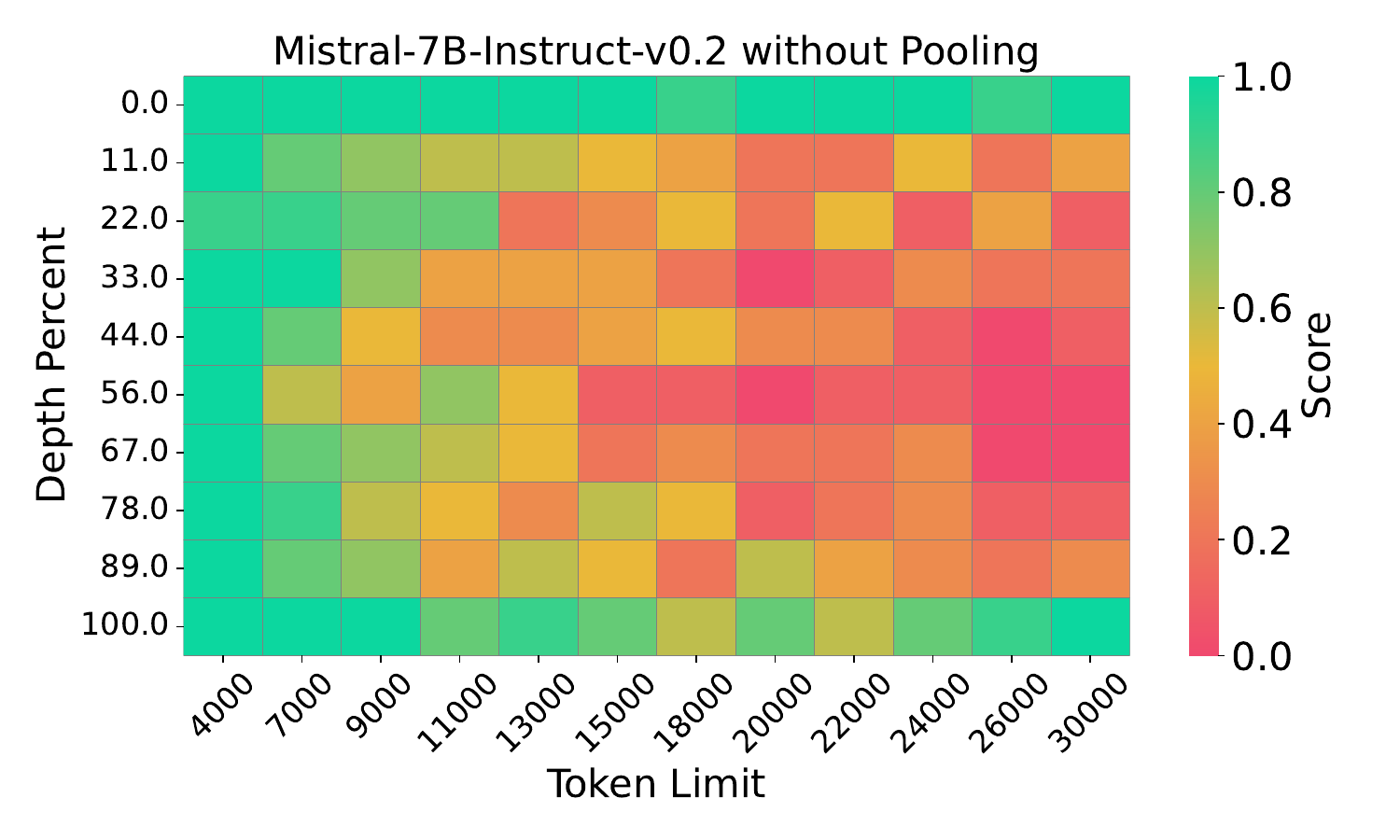}
    \includegraphics[width=0.49\textwidth]{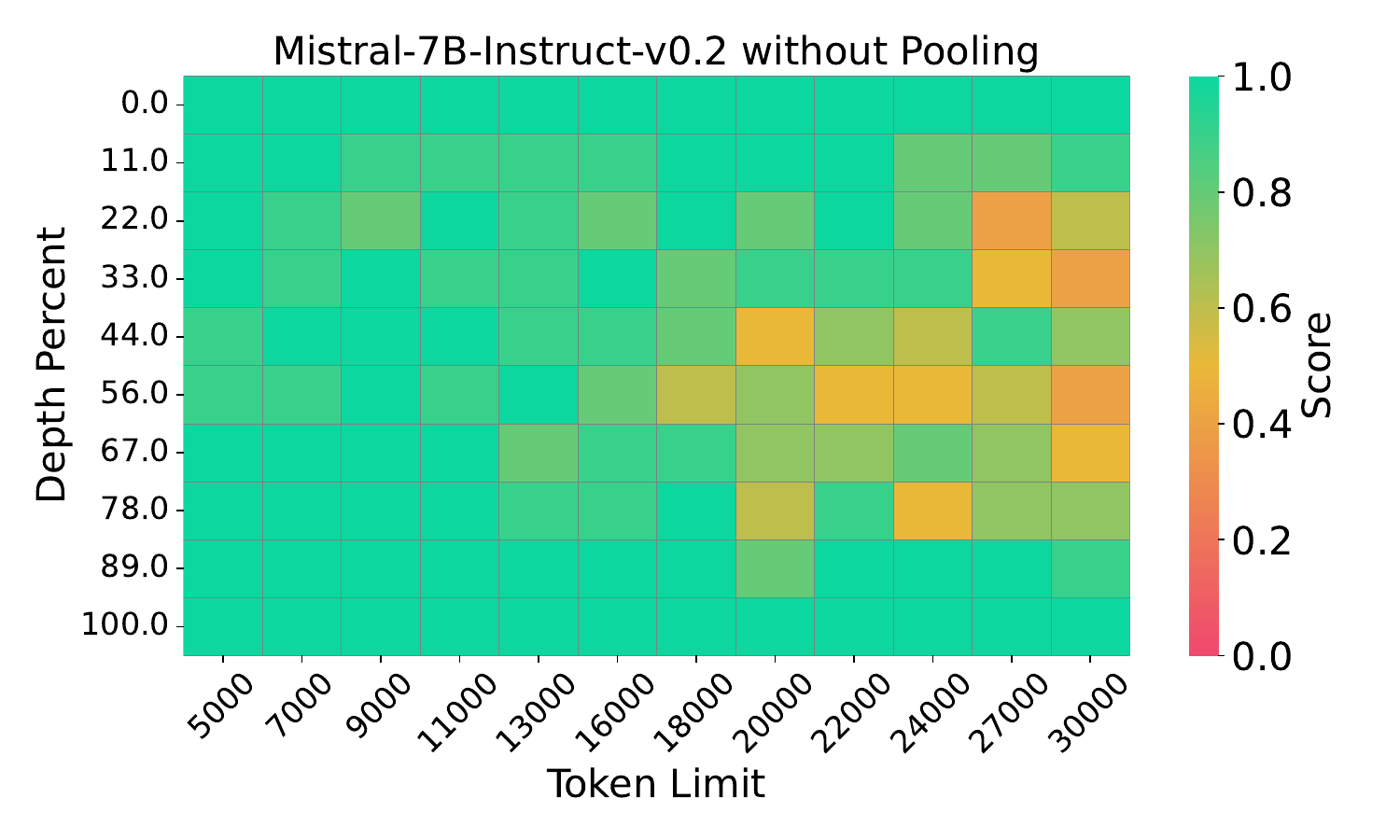}
    \vspace{-5pt}
    \caption{Ablation study of pooling on LongEval-Lines. The evaluation includes inputs, each comprised of lines formatted as "\texttt{line makeshift-penguin: REGISTER\_CONTENT is <10536>}", where the key is an adjective-noun pair and the value is a random 5-digit number. The model needs to retrieve the value based on a given key. 
    The x-axis denotes the length of the input; the y-axis indicates the position of the groundtruth, from 5K to 30K tokens. With the pooling, the model can retrieve correct values before 16k and performs significantly better than the one without pooling.}
    \vspace{-10pt}
    \label{fig: ablation}
\end{figure}

\vspace{-5pt}
\subsection{Ablation Study of Effectiveness of Pooling}\label{sec:ablation}
\vspace{-5pt}

We perform an ablation study on \texttt{Mistral-7B-Instruct-v0.2} to assess the impact of our pooling technique, a straightforward but efficient method for consolidating information through clustering. Our evaluation utilizes the modified LongEval-Lines benchmark~\cite{longchat2023}, incorporating randomly generated pairs and averaged scores. LongEval-Lines presents a greater challenge compared to Needle-in-a-Haystack because it involves identifying key-value pairs in noisy contexts of the same format, while in Needle-in-a-Haystack, the relevant information is more distinctly separated from other contexts. We apply max pooling with a kernel size of 5 and use the observation window with a size of 16, which are hyperparameters and could be customized according to different models. As illustrated in our results (Fig.~\ref{fig: ablation}), we find that pooling significantly enhances retrieval accuracy compared to methods not utilizing pooling. We hypothesize that this is because the initial portions of critical token clusters are weighted higher by attention mechanisms. Typically, large language models tend to copy the tokens surrounding the initial portions to keep the contextual integrity. However, naively compressed KV cache breaks this mechanism and could lead to partially correct results~(Fig.~\ref{fig: ablation}). Note that throughout our experiments, the choice between max pooling and average pooling did not yield significant differences in performance.

\begin{table*}[!t]

\fontsize{18}{24}\selectfont
\setlength{\tabcolsep}{5pt}
\centering
\caption{Performance comparison of \kv and H2O across various LLMs on LongBench.}\label{tab:longbench}
\begin{threeparttable}

\scalebox{0.3}{
\begin{tabular}{l|lcccccccccccccccc}
\specialrule{1pt}{0pt}{2pt}
&\multirow{4}{*}{~~~LLMs {\huge *}} & \multicolumn{3}{c}{Single-Document QA} & \multicolumn{3}{c}{Multi-Document QA}& \multicolumn{3}{c}{Summarization}& \multicolumn{3}{c}{Few-shot Learning}& \multicolumn{2}{c}{Synthetic} & \multicolumn{2}{c}{Code} \\
\cmidrule(lr){3-5}\cmidrule(lr){6-8}\cmidrule(lr){9-11}\cmidrule(lr){12-14}\cmidrule(lr){15-16}\cmidrule(lr){17-18}
&& \rotatebox[origin=c]{30}{NrtvQA} & \rotatebox[origin=c]{30}{Qasper} & \rotatebox[origin=c]{30}{MF-en} & \rotatebox[origin=c]{30}{HotpotQA} & \rotatebox[origin=c]{30}{2WikiMQA} & \rotatebox[origin=c]{30}{Musique} & \rotatebox[origin=c]{30}{GovReport} & \rotatebox[origin=c]{30}{QMSum} & \rotatebox[origin=c]{30}{MultiNews} & \rotatebox[origin=c]{30}{TREC} & \rotatebox[origin=c]{30}{TriviaQA} & \rotatebox[origin=c]{30}{SAMSum} & \rotatebox[origin=c]{30}{PCount} & \rotatebox[origin=c]{30}{PRe} & \rotatebox[origin=c]{30}{Lcc} & \rotatebox[origin=c]{30}{RB-P} \\

\specialrule{1pt}{2pt}{2pt}

\multirow{5}{*}{\rotatebox[origin=c]{90}{\fontsize{18}{100}\selectfont LWMChat}}

&\cellcolor{green!10}~~~All KV & \cellcolor{green!10}\textbf{18.18}&\cellcolor{green!10}\textbf{25.56}&\cellcolor{green!10} 40.94 &\cellcolor{green!10} 24.57 &\cellcolor{green!10} 19.39&\cellcolor{green!10} 10.49 &\cellcolor{green!10} \textbf{27.97} & \cellcolor{green!10}24.9 &\cellcolor{green!10} \textbf{24.81} &\cellcolor{green!10}71.0&\cellcolor{green!10} 60.9 & \cellcolor{green!10} 39.73 &\cellcolor{green!10} 3.17 &\cellcolor{green!10}3.5 & \cellcolor{green!10}44.4 & \cellcolor{green!10}43.82\\
\cline{2-18}

&\cellcolor{green!10}~~~\kv: 1024 & \cellcolor{green!10}18.02&\cellcolor{green!10}23.73&\cellcolor{green!10} 40.25 &\cellcolor{green!10} 24.61 &\cellcolor{green!10} \textbf{19.84}&\cellcolor{green!10} 10.77 &\cellcolor{green!10} 19.79 & \cellcolor{green!10}24.44 &\cellcolor{green!10} 23.53 &\cellcolor{green!10} 70.0 &\cellcolor{green!10} \textbf{61.42} & \cellcolor{green!10} 39.64 &\cellcolor{green!10} 1.67 &\cellcolor{green!10}3.0 & \cellcolor{green!10}43.34 & \cellcolor{green!10}44.0\\

&\cellcolor{green!10}~~~\kv: 2048 & \cellcolor{green!10}17.92&\cellcolor{green!10}25.03&\cellcolor{green!10} \textbf{41.38} &\cellcolor{green!10} 24.49 &\cellcolor{green!10} 19.38&\cellcolor{green!10} \textbf{11.34} &\cellcolor{green!10} 21.6 & \cellcolor{green!10}24.22 &\cellcolor{green!10} 24.36 &\cellcolor{green!10} 70.0 &\cellcolor{green!10} 61.11 & \cellcolor{green!10} 39.91 &\cellcolor{green!10} 2.17 &\cellcolor{green!10}4.0 & \cellcolor{green!10}44.46 & \cellcolor{green!10}\textbf{44.92}\\

&\cellcolor{green!10}~~~\kv: 4096 & \cellcolor{green!10}17.92&\cellcolor{green!10}25.47 &\cellcolor{green!10} 40.76 &\cellcolor{green!10} \textbf{24.92} &\cellcolor{green!10} 19.53&\cellcolor{green!10} 11.27 &\cellcolor{green!10} 25.34 & \cellcolor{green!10}\textbf{25.42} &\cellcolor{green!10} 24.58 &\cellcolor{green!10} 70.5 &\cellcolor{green!10} 61.08 & \cellcolor{green!10} 39.62 &\cellcolor{green!10} \textbf{3.17} &\cellcolor{green!10}\textbf{4.0} & \cellcolor{green!10}\textbf{44.49} & \cellcolor{green!10}44.08\\

&\cellcolor{green!10}~~~H2O: 4096 & \cellcolor{green!10}13.17&\cellcolor{green!10}24.82&\cellcolor{green!10}20.01&\cellcolor{green!10} 16.86 &\cellcolor{green!10} 9.74&\cellcolor{green!10} 7.2 &\cellcolor{green!10} 25.77 & \cellcolor{green!10}23.26 &\cellcolor{green!10} 23.83 &\cellcolor{green!10} \textbf{71.0} &\cellcolor{green!10} 61.06 & \cellcolor{green!10} \textbf{40.33} &\cellcolor{green!10} 0.0 &\cellcolor{green!10}0.0 & \cellcolor{green!10}41.52 & \cellcolor{green!10}40.97\\

\specialrule{1pt}{2pt}{10pt}\specialrule{1pt}{2pt}{2pt}

\multirow{5}{*}{\rotatebox[origin=c]{90}{\fontsize{18}{100}\selectfont LongChat}}

& \cellcolor{blue!10}~~~All KV & \cellcolor{blue!10}\textbf{20.88} & \cellcolor{blue!10}\textbf{29.36} & \cellcolor{blue!10}\textbf{43.2} & \cellcolor{blue!10}33.05 & \cellcolor{blue!10}24.58 & \cellcolor{blue!10}\textbf{14.66} & \cellcolor{blue!10}\textbf{30.89} & \cellcolor{blue!10}22.76 & \cellcolor{blue!10}\textbf{26.61} & \cellcolor{blue!10}\textbf{66.5} & \cellcolor{blue!10}\textbf{83.99} & \cellcolor{blue!10}\textbf{40.83} & \cellcolor{blue!10}0.0 & \cellcolor{blue!10}30.5 & \cellcolor{blue!10}54.89 & \cellcolor{blue!10}\textbf{59.05} \\
\cline{2-18}

& \cellcolor{blue!10}~~~\kv: 1024 & \cellcolor{blue!10}19.32 & \cellcolor{blue!10}26.6 & \cellcolor{blue!10}37.93 & \cellcolor{blue!10}34.15 & \cellcolor{blue!10}23.34 & \cellcolor{blue!10}12.71 & \cellcolor{blue!10}23.45 & \cellcolor{blue!10}21.81 & \cellcolor{blue!10}24.93 & \cellcolor{blue!10}65.0 & \cellcolor{blue!10}80.88 & \cellcolor{blue!10}38.19 & \cellcolor{blue!10}0.0 & \cellcolor{blue!10}31.0 & \cellcolor{blue!10}53.63 & \cellcolor{blue!10}57.62 \\

& \cellcolor{blue!10}~~~\kv: 2048 & \cellcolor{blue!10}19.28 & \cellcolor{blue!10}28.81 & \cellcolor{blue!10}40.26 & \cellcolor{blue!10}\textbf{35.31} & \cellcolor{blue!10}23.75 & \cellcolor{blue!10}13.44 & \cellcolor{blue!10}26.3 & \cellcolor{blue!10}22.29 & \cellcolor{blue!10}25.73 & \cellcolor{blue!10}66.0 & \cellcolor{blue!10}79.93 & \cellcolor{blue!10}39.59 & \cellcolor{blue!10}0.0 & \cellcolor{blue!10}\textbf{31.0} & \cellcolor{blue!10}\textbf{56.05} & \cellcolor{blue!10}58.61 \\

& \cellcolor{blue!10}~~~\kv: 4096 & \cellcolor{blue!10}20.68 & \cellcolor{blue!10}29.34 & \cellcolor{blue!10}42.21 & \cellcolor{blue!10}33.95 & \cellcolor{blue!10}\textbf{24.88} & \cellcolor{blue!10}14.15& \cellcolor{blue!10}28.55 & \cellcolor{blue!10}\textbf{23.11} & \cellcolor{blue!10}26.45 & \cellcolor{blue!10}66.0 & \cellcolor{blue!10}\textbf{81.25} & \cellcolor{blue!10}\textbf{40.52} & \cellcolor{blue!10}0.0 & \cellcolor{blue!10}29.5 & \cellcolor{blue!10}54.79 & \cellcolor{blue!10}58.81 \\

& \cellcolor{blue!10}~~~H2O: 4096 & \cellcolor{blue!10}19.31 & \cellcolor{blue!10}28.3 & \cellcolor{blue!10}37.75 & \cellcolor{blue!10}30.51 & \cellcolor{blue!10}23.06 & \cellcolor{blue!10}11.76 & \cellcolor{blue!10}27.55 & \cellcolor{blue!10}21.37 & \cellcolor{blue!10}26.49 & \cellcolor{blue!10}66.0 & \cellcolor{blue!10}75.8 & \cellcolor{blue!10}39.92 & \cellcolor{blue!10}0.0 & \cellcolor{blue!10}25.5 & \cellcolor{blue!10}53.56 & \cellcolor{blue!10}55.53 \\

\specialrule{1pt}{2pt}{10pt}\specialrule{1pt}{2pt}{2pt}

\multirow{5}{*}{\rotatebox[origin=c]{90}{\fontsize{18}{100}\selectfont Mistral}}

& \cellcolor{red!10}~~~All KV & \cellcolor{red!10}\textbf{26.82} & \cellcolor{red!10}33.06 & \cellcolor{red!10}49.28 & \cellcolor{red!10}\textbf{42.77} & \cellcolor{red!10}27.33 & \cellcolor{red!10}19.27 & \cellcolor{red!10}\textbf{32.85} & \cellcolor{red!10}24.25 & \cellcolor{red!10}27.06 & \cellcolor{red!10}71.0 & \cellcolor{red!10}86.23 & \cellcolor{red!10}42.98 & \cellcolor{red!10}2.75 & \cellcolor{red!10}86.98 & \cellcolor{red!10}55.51 & \cellcolor{red!10}\textbf{52.88} \\
\cline{2-18}

& \cellcolor{red!10}~~~\kv: 1024 & \cellcolor{red!10}25.54 & \cellcolor{red!10}29.51 & \cellcolor{red!10}49.25 & \cellcolor{red!10}40.94 & \cellcolor{red!10}25.7 & \cellcolor{red!10}\textbf{19.42} & \cellcolor{red!10}25.89 & \cellcolor{red!10}23.82 & \cellcolor{red!10}26.11 & \cellcolor{red!10}69.5 & \cellcolor{red!10}\textbf{86.48} & \cellcolor{red!10}42.06 & \cellcolor{red!10}2.98 & \cellcolor{red!10}\textbf{88.56} & \cellcolor{red!10}55.65 & \cellcolor{red!10}51.87 \\

& \cellcolor{red!10}~~~\kv: 2048 & \cellcolor{red!10}25.89 & \cellcolor{red!10}32.47 & \cellcolor{red!10}48.6 & \cellcolor{red!10}41.71 & \cellcolor{red!10}27.31 & \cellcolor{red!10}18.69 & \cellcolor{red!10}28.81 & \cellcolor{red!10}\textbf{24.5} & \cellcolor{red!10}26.6 & \cellcolor{red!10}70.0 & \cellcolor{red!10}86.27 & \cellcolor{red!10}42.47 & \cellcolor{red!10}3.09 & \cellcolor{red!10}87.43 & \cellcolor{red!10}\textbf{55.93} & \cellcolor{red!10}52.01 \\

& \cellcolor{red!10}~~~\kv: 4096 & \cellcolor{red!10}26.41 & \cellcolor{red!10}\textbf{33.36} & \cellcolor{red!10}\textbf{49.81} & \cellcolor{red!10}42.32 & \cellcolor{red!10}\textbf{27.93} & \cellcolor{red!10}18.76 & \cellcolor{red!10}30.74 & \cellcolor{red!10}24.19 & \cellcolor{red!10}\textbf{27.08} & \cellcolor{red!10}\textbf{71.0} & \cellcolor{red!10}86.25 & \cellcolor{red!10}\textbf{43.01} & \cellcolor{red!10}2.73 & \cellcolor{red!10}86.18 & \cellcolor{red!10}55.62 & \cellcolor{red!10}52.65 \\

& \cellcolor{red!10}~~~H2O: 4096 & \cellcolor{red!10}22.61 & \cellcolor{red!10}29.06 & \cellcolor{red!10}47.22 & \cellcolor{red!10}36.54 & \cellcolor{red!10}20.6 & \cellcolor{red!10}16.25 & \cellcolor{red!10}30.0 & \cellcolor{red!10}23.8 & \cellcolor{red!10}26.75 & \cellcolor{red!10}70.5 & \cellcolor{red!10}86.16 & \cellcolor{red!10}42.97 & \cellcolor{red!10}\textbf{3.46} & \cellcolor{red!10}86.38 & \cellcolor{red!10}53.72 & \cellcolor{red!10}51.1 \\

\specialrule{1pt}{2pt}{10pt}\specialrule{1pt}{2pt}{2pt}

\multirow{5}{*}{\rotatebox[origin=c]{90}{\fontsize{18}{100}\selectfont Mixtral}}

& \cellcolor{cyan!10}~~~All KV & \cellcolor{cyan!10}{26.81} & \cellcolor{cyan!10}\textbf{37.06} & \cellcolor{cyan!10}51.55 & \cellcolor{cyan!10}47.77 & \cellcolor{cyan!10}32.46 & \cellcolor{cyan!10}\textbf{26.59} & \cellcolor{cyan!10}\textbf{34.25} & \cellcolor{cyan!10}\textbf{26.05} & \cellcolor{cyan!10}27.91 & \cellcolor{cyan!10}76.0 & \cellcolor{cyan!10}90.57 & \cellcolor{cyan!10}46.98 & \cellcolor{cyan!10}5.5 & \cellcolor{cyan!10}100.0 & \cellcolor{cyan!10}\textbf{69.07} & \cellcolor{cyan!10}69.65 \\
\cline{2-18}

& \cellcolor{cyan!10}~~~\kv: 1024 & \cellcolor{cyan!10}26.01 & \cellcolor{cyan!10}34.65 & \cellcolor{cyan!10}51.58 & \cellcolor{cyan!10}\textbf{48.23} & \cellcolor{cyan!10}32.67 & \cellcolor{cyan!10}25.92 & \cellcolor{cyan!10}27.77 & \cellcolor{cyan!10}25.0 & \cellcolor{cyan!10}27.25 & \cellcolor{cyan!10}74.5 & \cellcolor{cyan!10}90.42 & \cellcolor{cyan!10}46.48 & \cellcolor{cyan!10}5.5 & \cellcolor{cyan!10}99.5 & \cellcolor{cyan!10}69.02 & \cellcolor{cyan!10}68.98 \\

& \cellcolor{cyan!10}~~~\kv: 2048 & \cellcolor{cyan!10}\textbf{27.12} & \cellcolor{cyan!10}36.9 & \cellcolor{cyan!10}51.91 & \cellcolor{cyan!10}47.46 & \cellcolor{cyan!10}33.23 & \cellcolor{cyan!10}26.27 & \cellcolor{cyan!10}30.19 & \cellcolor{cyan!10}25.84 & \cellcolor{cyan!10}27.8 & \cellcolor{cyan!10}\textbf{76.0} & \cellcolor{cyan!10}90.24 & \cellcolor{cyan!10}46.31 & \cellcolor{cyan!10}5.5 & \cellcolor{cyan!10}100.0 & \cellcolor{cyan!10}68.72 & \cellcolor{cyan!10}\textbf{70.01} \\

& \cellcolor{cyan!10}~~~\kv: 4096 & \cellcolor{cyan!10}26.46 & \cellcolor{cyan!10}37.03 & \cellcolor{cyan!10}\textbf{52.62} & \cellcolor{cyan!10}47.71 & \cellcolor{cyan!10}\textbf{33.35} & \cellcolor{cyan!10}26.45 & \cellcolor{cyan!10}32.64 & \cellcolor{cyan!10}25.87 & \cellcolor{cyan!10}\textbf{27.94} & \cellcolor{cyan!10}75.5 & \cellcolor{cyan!10}\textbf{90.71} & \cellcolor{cyan!10}\textbf{47.14} & \cellcolor{cyan!10}5.5 & \cellcolor{cyan!10}\textbf{100.0} & \cellcolor{cyan!10}68.81 & \cellcolor{cyan!10}69.56 \\

& \cellcolor{cyan!10}~~~H2O: 4096 & \cellcolor{cyan!10}20.45 & \cellcolor{cyan!10}32.09 & \cellcolor{cyan!10}48.02 & \cellcolor{cyan!10}34.76 & \cellcolor{cyan!10}25.69 & \cellcolor{cyan!10}16.5 & \cellcolor{cyan!10}29.76 & \cellcolor{cyan!10}23.53 & \cellcolor{cyan!10}26.84 & \cellcolor{cyan!10}74.5 & \cellcolor{cyan!10}90.24 & \cellcolor{cyan!10}47.1 & \cellcolor{cyan!10}\textbf{7.06} & \cellcolor{cyan!10}99.42 & \cellcolor{cyan!10}64.91 & \cellcolor{cyan!10}63.52 \\
\specialrule{1pt}{2pt}{0pt}
\end{tabular}
}

\begin{tablenotes}
    \scriptsize
    \item[] 
    \hspace{-20pt}{\scriptsize *} Credit to Jin et al.~\cite{jin2024llm} for the template used in the table.
\end{tablenotes}
\end{threeparttable}\vspace{-10pt}

\end{table*}

\vspace{-5pt}
\subsection{Experiments on LongBench}
\vspace{-5pt}

We evaluate \kv on these four models using LongBench~\cite{bai2023longbench}, a multi-task benchmark designed to rigorously evaluate long context understanding capabilities across various datasets, spanning single and multi-document QA, summarization, few-shot learning, synthetic tasks, and code completion. We choose \texttt{LWM-Text-Chat-1M} with 1 million context length, \texttt{LongChat-7b-v1.5-32k}, \texttt{Mistral-7B-Instruct-v0.2}, \texttt{Mixtral-8x7B-Instruct-v0.1} with 32k context length as our baselines. For each model, we test \kv with various settings: compressing KV caches in the prompt to 1024, 2048, and 4096 tokens. We use max pooling with kernel size 7 and observation window size 32. Table \ref{tab:longbench} illustrates a negligible performance drop from models with \kv compared with original implementations for 16 different datasets, even with prompt-KV with 1024 tokens. Some models even outperform the baseline. Our results substantiate that \kv can grasp the key information in the long context and give comprehensive summaries with details. Moreover, our results also indicate the effectiveness of \kv in compressing the prompt KV cache. 
For these 4 models, the average input token length is around 13k. Thus, using 1024, \kv achieves an average compression rate of 92\%, and using 4096, it reaches 68\%, all with negligible drops in accuracy.
We compare \kv and H2O on the LongBench dataset to further demonstrate the performance of \kv. To fairly evaluate the accuracy, we set the prompt capacity for H2O to 4096. As Table~\ref{tab:longbench} shows, \kv delivers significantly better performance than H2O. Even with 1024 prompt KV caches, \kv on \texttt{Mistral-7B-Instruct-v0.2} achieves better performance than H2O with 4096 caches on 11 out of 16 benchmarks. 
\subsection{Experiments on Command-R}
To further assess the performance of \kv, we conduct experiments using Cohere's \texttt{Command-R} model \cite{coherecommandr}, an open-source model with 35B parameters and capable of handling sequences of up to 128k token length. \texttt{Command-R} is designed for complex tasks requiring long context, such as retrieval-augmented generation (RAG). We extensively test \texttt{Command-R} on NarrativeQA and a modified version of the Needle-in-a-Haystack where it achieves promising results. To evaluate \kv's impact on RAG, we ran tests on bioasq \cite{bioasq}, multi-hop question answering with HotpotQA \cite{yang2018hotpotqa}, and an internal benchmark on tool use, which further demonstrated its effectiveness. Throughout all experiments, we limit the KV cache to a maximum of 4096 tokens, while the pooling kernel size and window size are set to 13 and 64, respectively. For our evaluations, these hyper-parameters give a KV cache compression ratio between 2x to 32x depending on the sequence length.

\subsubsection{Needle-in-a-Haystack}
In previous experiments \cite{anthropic2023needlesprompt}, it was noted that Needle-in-a-Haystack \cite{kamradt2023needle} evaluation was heavily influenced by the specific context used. To address this issue, we modify the evaluation by permuting context compositions for each length and depth combination. This approach, which we ran eight times, yielded more robust results. We observe a slight decrease in scores across all models tested under this setting compared to the original setup with no context shuffling. For simplicity, we aggregated the scores across all depths and lengths for the baseline model and the one with \kv. As seen in Table \ref{table:commandr_needles}, applying \kv to \texttt{Command-R} shows no degradation in performance, even with a 128k sequence length resulting in 32x compression of KV cache.

\begin{table}[h]
\centering
\caption{Needles-in-a-Haystack Test Results}
\begin{tabular}{lccc}
\toprule
\text{Model} & \texttt{Command-R} & \texttt{Command-R} + \kv & \text{\% Difference} \\
\midrule
Score & 9.866 & 9.819 & -0.5\% \\
\bottomrule
\end{tabular}
\label{table:commandr_needles}
\end{table}

\subsubsection{Retrieval Augmented Generation (RAG)}
We assess \kv's effectiveness in RAG tasks, which are more intricate than synthetic long-context tasks like Needle-in-a-Haystack and closer to real use cases compared to tasks like NarrativeQA. RAG tasks require selecting pertinent documents from an indexed corpus based on the given prompt. An expanded context window enables the retrieval of additional documents, which can lead to improved model performance. However, this also increases memory requirements and latency, highlighting the delicate balance between retrieval scope and system resources. \kv proves beneficial in these tasks by reducing memory usage while enhancing the performance. We evaluated \kv's impact on RAG tasks with sequence lengths up to approximately 40,000 tokens.

\paragraph{RAG Citation}
We begin by assessing \kv's impact on the model's ability to select relevant documents, a crucial aspect of effective RAG. We evaluate on an internal benchmarks from Cohere. The setup of the benchmark is as follow: for each prompt, we gathered a set of topic-related documents that included ground truth answers along with a sample of negative documents ensuring a total of 100 documents per prompt. We measured the model's performance by calculating the F1-score when the model successfully retrieved the ground truth documents. The dataset employed in this experiment spanned context lengths from 20,000 to 40,000 tokens. Given our KV cache size of 4096, we achieve a compression of 5-10x. As observed in Table \ref{table:commandr_cite}, \kv demonstrates a remarkable ability to retain nearly 98.8\% of \texttt{Command-R}'s performance.


\begin{table}[h]
\centering
\caption{RAG Test Results}
\begin{tabular}{lcc}
\toprule
\text{Evaluation Task} & \text{Metric} & \text{\% Difference} \\
\midrule
RAG Citation & F1 score & -1.2\% \\
RAG End-to-end & F1 score & -2.1\% \\
\bottomrule
\end{tabular}
\label{table:commandr_cite}
\end{table}


\paragraph{Generation} \label{Generation_Section}

          

As the quality of generation is important to a model's RAG capability, we evaluate \texttt{Command-R} on lost-in-the-middle and generation quality. Lost-in-the-middle is aimed to analyze whether the performance of the model varies when altering the position of ground-truth information in the context \cite{lostinthemiddle}. The latter is a relatively simple metric where we define the accuracy of the model to be the proportion of the ground-truth answer phrase appearing in model's response. We conducted 3 experiments with 30, 100 and 200 sampled documents for each ground-truth. We repeat each experiment 3 times and insert the relevant documents at beginning, middle and end of the context to test \kv's robustness.We report the relative difference to the baseline model. The dataset used in this phase is based on the bioasq dataset \cite{bioasq} with RAG-style formulation from Cohere \cite{cohererag}.

\begin{table}[ht]
\centering
\caption{RAG Generation Test Results on bioasq}
\label{table:commandr_generation}
\begin{tabular}{@{}cccc@{}}
\toprule
Number of Documents & Approximate Context Length & Ground Truth Position & \% Difference \\ \midrule
\multirow{4}{*}{30}  & \multirow{4}{*}{8k}  & 0  & -1.8\% \\
                     &                      & 14 & 0\%   \\
                     &                      & 30 & -3.4\% \\
                     &                      & Avg & -1.7\% \\ \midrule
\multirow{4}{*}{100} & \multirow{4}{*}{14k} & 0  & -1.2\% \\
                     &                      & 14 & +0.9\% \\
                     &                      & 30 & -0.9\% \\
                     &                      & Avg & -0.6\% \\ \midrule
\multirow{4}{*}{200} & \multirow{4}{*}{24k} & 0  & +4.9\% \\
                     &                      & 14 & +4.9\% \\
                     &                      & 30 & +6.4\% \\
                     &                      & Avg & +5.4\% \\ \bottomrule
\end{tabular}
\smallskip
\small
\textit{Note:} For each number of sampled documents, we report the approximate context length and the difference from the baseline at each ground-truth position.
\end{table}


As Table \ref{table:commandr_generation} shows, \kv is robust in terms of generation quality and does not suffer from the well-known lost-in-the-middle pathology. Moreover, \kv improves performance over the baseline model when the context contains close to 200 documents. One potential explanation to this is that by adequately compressing the KV cache, we can effectively reduce the noise from negative documents and push the model to construct attention scores more focused on the relevant information.

\paragraph{End-to-End RAG} 
To assess \kv's robustness in a comprehensive manner, we integrated it into a complete RAG pipeline. This evaluation starts by retrieving 200 documents using Cohere's embedding service \cite{cohereembed} in response to a given query. These documents were then re-ranked using Cohere's re-ranking model \cite{coherererank}, which filtered out half of the candidates, resulting in a list of 100 documents. We prompt \texttt{Command-R} using this list and calculate the accuracy metric as described in Section \ref{Generation_Section}. We employed a modified version of the HotpotQA dataset \cite{yang2018hotpotqa} and leveraged Wikipedia as the document source. This setup introduces a more challenging set of documents as all documents, relevant or not, are semantically similar.



Table \ref{table:commandr_cite} showcases \kv's robust performance in a production-like RAG setting. With an average dataset length of around 16,000 tokens, the KV cache benefits from a compression ratio of approximately 4x.

\begin{figure}[ht]
    \centering
    \includegraphics[width=0.8\textwidth]{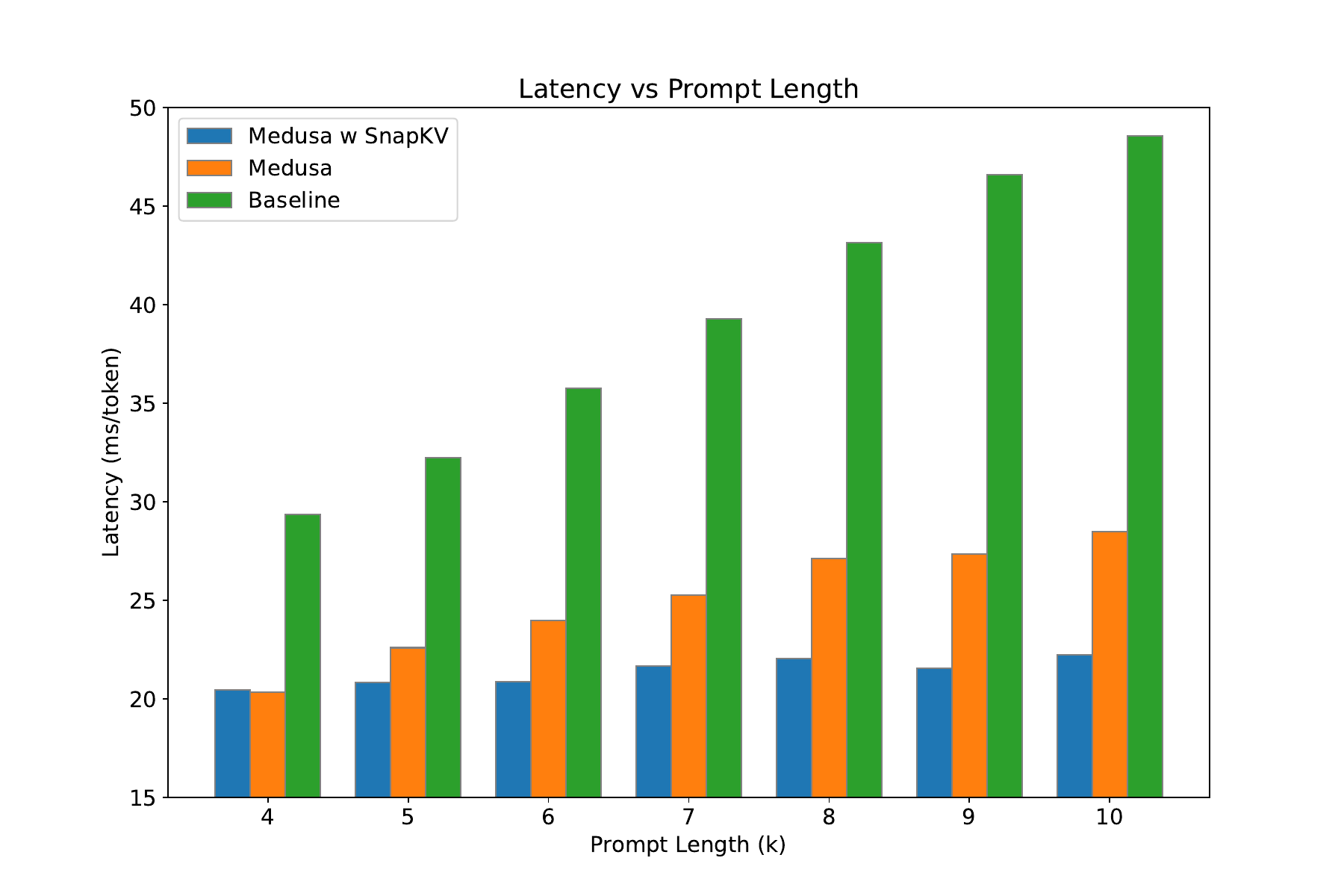}
    \caption{Comparison of generation speed (ms/token). The baseline is the Huggingface implementation of naive decoding.}
    \label{fig: medusa}
\end{figure}
\subsection{Case Study: Compatibility with Parallel Decoding}

In this section, we provide a novel perspective on employing KV cache compression synergistically with parallel decoding~\cite{stern2018blockwise, leviathan2023fast, chen2023accelerating, miao2023specinfer, zhang2024recurrent}. Parallel decoding leverages a lightweight model or an adaptor to draft initial tokens, which are subsequently verified by larger LLMs. This strategy effectively reduces memory overhead, a critical concern given the autoregressive nature of LLMs that renders them more memory-intensive than computationally demanding. Specifically, in LLMs, each decoding step involves generating a single token, with the transfer of weights between High Bandwidth Memory (HBM) and cache contributing to significant overhead~\cite{dao2022flashattention,dao2023flashattention}.

Our investigation incorporates \kv with \texttt{Medusa}~\cite{cai2024medusa}\footnote{\url{https://github.com/FasterDecoding/Medusa}}, a cutting-edge parallel decoding framework that utilizes multiple classifiers and tree attention mechanisms for drafting tokens, subsequently verified by LLMs. One of the challenges identified is the issue of speculative decoding in processing long sequences since generating multiple tokens per decoding step introduces computational bottlenecks during long sequence processing, such as query-key matrix multiplication tiling~\cite{dao2023flash}. By maintaining a constant size for the KV cache associated with prompts during generation, \kv enhances generation efficiency.

Empirical results shown in Figure~\ref{fig: medusa} highlight the performance across various prompt lengths, with \texttt{Mistral-7B-Instruct-v0.2}\footnote{\href{https://huggingface.co/text-generation-inference/Mistral-7B-Instruct-v0.2-medusa/tree/main}{TGI trained \texttt{Medusa} heads}} undergoing a maximum of 128 generation steps unless preemptively halted. The experiments utilized a subset of the QASPER~\cite{dasigi2021dataset}, with a fixed prompt instructing the LLM to summarize the paper. The truncation strategy adopted aligns with LongBench~\cite{bai2023longbench} standards, by removing the context in the middle to achieve the desired sequence length for benchmarking.

The findings indicate a slowdown in \texttt{Medusa}'s performance as sequence lengths extend, a challenge effectively mitigated by \kv's intervention, which achieved a 1.3x speedup for sequences with 10k length compared to \texttt{Medusa} and a 2.2x speedup compared to the native decoding. This improvement underscores the potential of combining KV cache compression with parallel decoding frameworks to enhance LLM efficiency, particularly in long-context scenarios.
\vspace{-5pt}
\section{Discussions}
\vspace{-5pt}
\label{sec: conclusion}

\kv is an effective yet straightforward solution that compresses the KV cache to mitigate the computational and memory burdens of processing extensive prompts. Observing that specific tokens within prompts gain consistent attention from each head during generation, our methodology not only retrieve crucial information but also enhances processing efficiency. 
Despite its strengths, \kv's scope is primarily confined to the generative aspect of models, specifically targeting the KV cache during the generation. This limitation implies that \kv cannot extend a model's long context capability if the model inherently struggles with long contexts or exhibits poor performance. Additionally, \kv's design does not cover the processing of the prompt inference, which limits its effectiveness in scenarios where the system cannot handle prompts of extensive length.
Nonetheless, our contributions offer significant insights and tools for the community, paving the way for more refined approaches on managing the challenges of large-scale language modeling. The appendix provides more experiments with parallel decoding and the discussion about generation speedup.

\bibliographystyle{unsrtnat}
\bibliography{ref.bib}

\newpage
\appendix
\appendix

\section{Discussion of Generation Time Speedup}
\begin{figure}[ht]
    \centering
    \includegraphics[width=0.9\textwidth]{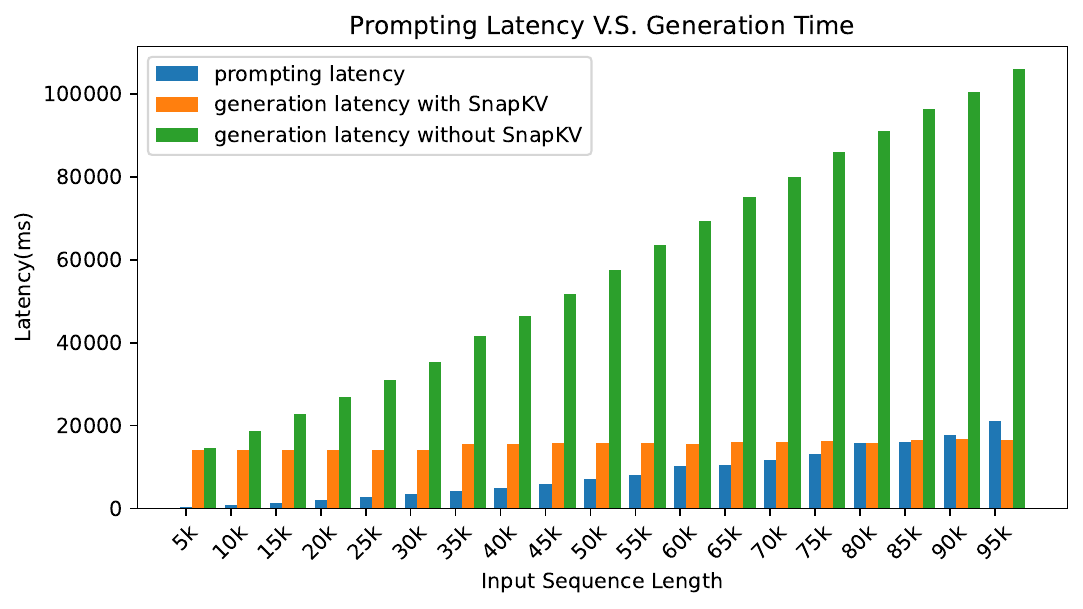}
    \caption{
    The prompting time and generation time comparison between Mistral model with and without SnapKV.
    }
    \label{fig: snapkv_timebreakdown}
\end{figure}

To better assess \kv's effectiveness across different stages, we documented a detailed time breakdown for \texttt{Mistral-7B-Instruct-v0.2} during both the prompting and generation stages. We configured the model to consistently generate 512 tokens, facilitating a direct comparison with the prompting stage. As illustrated in Figure \ref{fig: snapkv_timebreakdown}, generation time dominates the whole processing time for LLMs over input sequences, introducing significant overhead. While the generation time for the original model increases with input length, \kv maintains a consistent decoding speed regardless of input length, significantly reducing generation time. Especially, \kv is able to achieve balanced prompting time and generation time with input length smaller than 100k. 

\newpage

\section{Visulization of the Generated Context}
\vspace{-5pt}

\begin{figure}[ht]
\centering
\vspace{-2mm}
\includegraphics[width=\linewidth]
{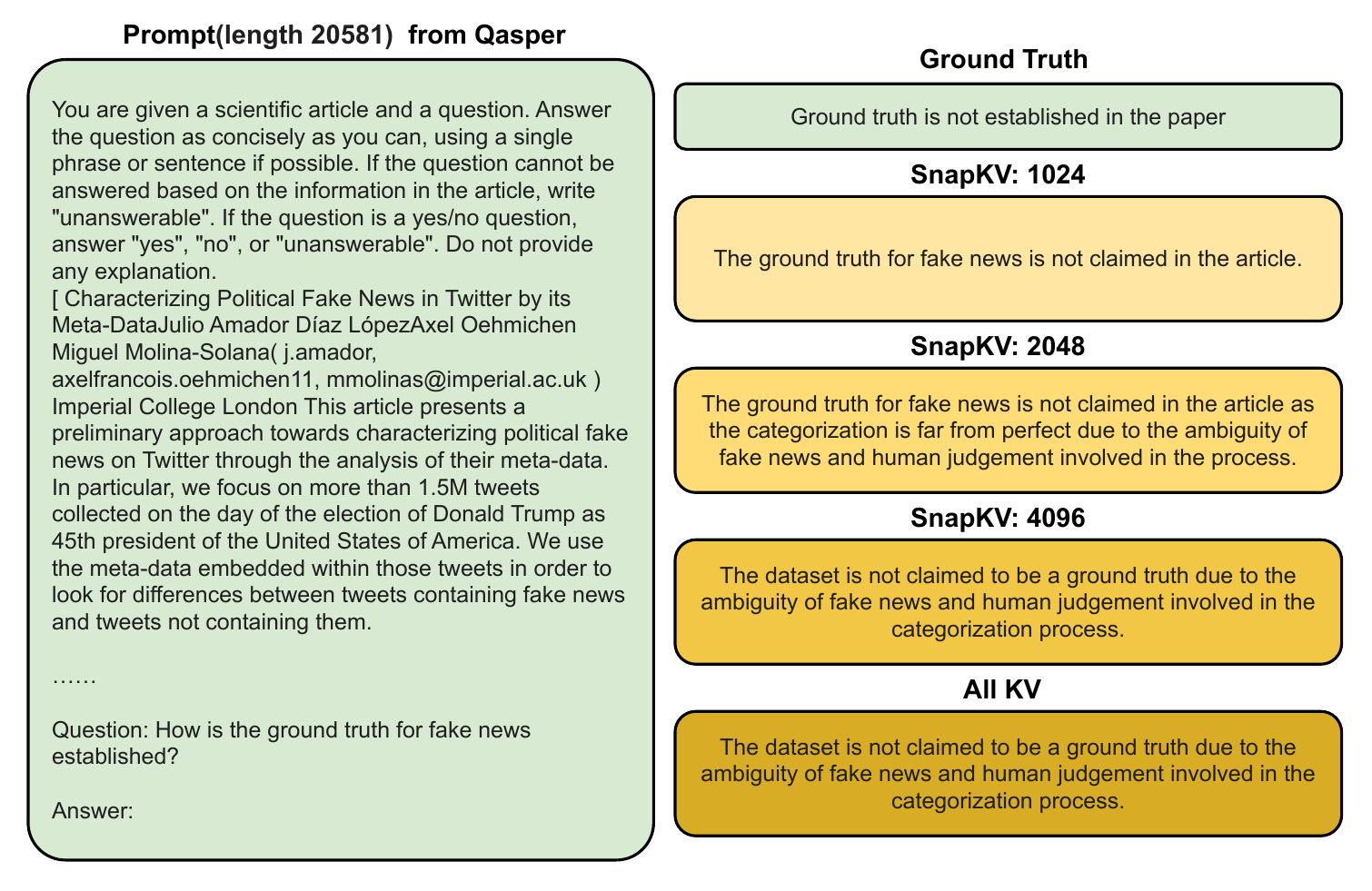}
\includegraphics[width=\linewidth]
{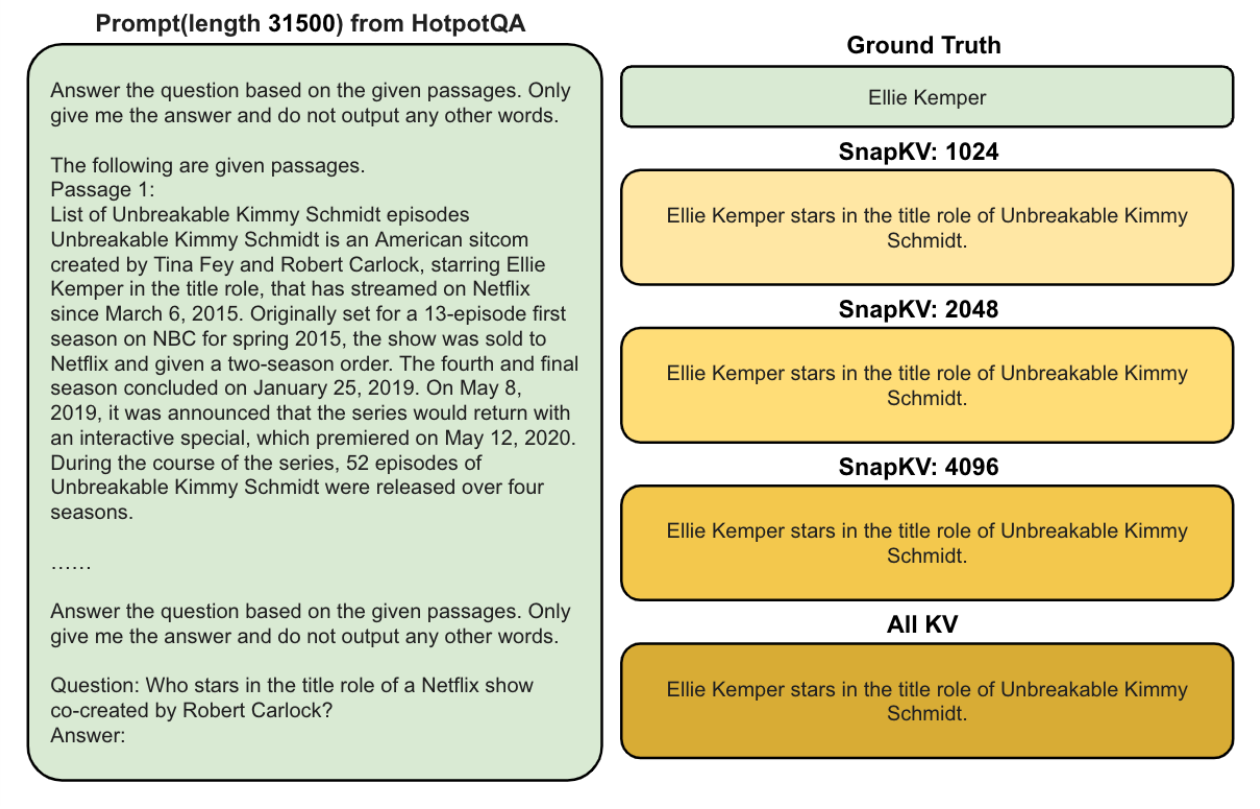}
\caption{Visualization of generation examples from Samsum, Qasper, HotpotQA datasets with \texttt{mistral-7B-instruct-v0.2}. Results are compared between ground truth, \kv with 1024 prompt tokens, with 2048, with 4096, the baseline model with full KV cache.}
\label{fig:generated_text1}
\end{figure}

\newpage

\end{document}